\documentclass{IEEEtran}
\IEEEoverridecommandlockouts

\usepackage{cite}
\usepackage{amsmath,amssymb,amsfonts}
\usepackage{algorithmic}
\usepackage{graphicx}
\usepackage{textcomp}
\usepackage{xcolor}
\usepackage{xargs}
\usepackage{url}
\newcommand{\citeb}[1]{\cite{#1}}

\usepackage{booktabs}
\usepackage{tabularx}

\usepackage[roman]{parnotes}
\makeatletter
\def\parnoteclear{%
    \gdef\PN@text{}%
    \parnotereset
}
\makeatother

\usepackage{mathtools}
\DeclarePairedDelimiter\abs{\lvert}{\rvert}

\newcommand{\mx}[1]{\mathbf{#1}}

\usepackage{siunitx}

\newcommand{\ms}[1]{\SI{#1}{\milli\second}}

\usepackage{tablefootnote}

\usepackage[nomessages]{fp}
\newcommand{\calc}[2]{\FPeval{\calcresult}{round(#2,#1)}\calcresult}

\usepackage{multirow}
\newcolumntype{H}{>{\setbox0=\hbox\bgroup}c<{\egroup}@{}} % hide col
\usepackage{pgfplots}
\usepackage{pgfplotstable}
\usepgfplotslibrary{groupplots}
\pgfplotsset{compat=newest}
\newcommand{\figref}[1]{Figure~\ref{#1}} 
\newcommand{\tblref}[1]{Table~\ref{#1}} 
\newcommand{\secref}[1]{Section~\ref{#1}} 
\graphicspath{{./figs/},{./}} %arXiv compatible
\newcommand{\plotdataDir}[1]{./plotdata-merged/#1}

\newcommand{\plotdataDirPfm}[1]{./plotdata-merged/#1} 

\newcommand{\change}[1]{#1}
\newcommand{\revision}[1]{#1}
\newcommand{\revisionB}[1]{#1}

\pgfplotscreateplotcyclelist{seqStyleList}{
    {blue!50!white,mark=diamond,mark size=1.5pt},
    {brown!70!black,mark=otimes,mark size=1.5pt},
    {red!90!black,mark=square,mark size=1.5pt},
    {orange,mark=star,mark size=1.5pt},
    {blue!80!black,mark=x,mark size=1.5pt},
    {green!80!black,mark=o,mark size=1.5pt},
    {green!80!white,mark=*,mark size=1.5pt},
}
\pgfplotscreateplotcyclelist{seqColorList}{
    {blue!50!white,mark size=1.5pt},
    {brown!70!black,mark size=1.5pt},
    {red!90!black,mark size=1.5pt},
    {orange,mark size=1.5pt},
    {blue!80!black,mark size=1.5pt},
    {green!80!black,mark size=1.5pt},
}

\begin{document}
\title{\huge
CBinfer: Exploiting Frame-to-Frame Locality for Faster Convolutional Network Inference on Video Streams}

\author{Lukas Cavigelli, Luca Benini
\thanks{The authors would like to thank \emph{armasuisse Science \& Technology} for funding this research. This project was supported in part by the EU's H2020 programme under grant no. 732631 (OPRECOMP).

L. Cavigelli and L. Benini are with ETH Z\"urich, Z\"urich, Switzerland (e-mail: \{cavigelli,benini\}@iis.ee.ethz.ch). }}
\maketitle
\begin{abstract}
\change{
The last few years have brought advances in computer vision at an amazing pace, grounded on new findings in deep neural network construction and training as well as the availability of large labeled datasets. Applying these networks to images demands a high computational effort and pushes the use of state-of-the-art networks on real-time video data out of reach of embedded platforms. 

Many recent works focus on reducing network complexity for real-time inference on embedded computing platforms. We adopt an orthogonal viewpoint and propose a novel algorithm exploiting the spatio-temporal sparsity of pixel changes. This optimized inference procedure resulted in an average speed-up of $9.1\times$ over cuDNN on the Tegra~X2 platform at a negligible accuracy loss of $<0.1\%$ and no retraining of the network for a semantic segmentation application. Similarly, an average speed-up of $7.0\times$ has been achieved for a pose detection DNN and a reduction of $5\times$ of the number of arithmetic operations to be performed for object detection on static camera video surveillance data. These throughput gains combined with a lower power consumption result in an energy efficiency of 511\,GOp/s/W compared to 70\,GOp/s/W for the baseline.}
\end{abstract}

\section{Introduction}
Computer vision (CV) technology has become a key ingredient for automatized data analysis over a broad range of real-world applications: smart cameras for video surveillance, robotics, industrial quality assurance, medical diagnostics, and advanced driver assistance systems have recently become popular due the rising accuracy and robustness of CV algorithms. 
This industry interest has fostered the procedure of a wealth of research projects yielding a fierce competition on many benchmarks datasets such as the ImageNet/ILSVRC \citeb{Russakovsky2014}, MS COCO \citeb{Lin2014}, and Cityscapes \citeb{Cordts2016} benchmarks, on which scientists from academia and big industry players evaluate their latest algorithms. 

In recent years, the most competitive approaches to address many CV challenges have relied on machine learning with complex, multi-layered, trained feature extractors commonly referred to as \emph{deep learning} \citeb{Goodfellow-et-al-2016}. 
The most frequently used flavor of deep learning techniques for CV are convolutional neural networks (ConvNets, CNNs). Since their landslide success at the 2012 ILSVRC competition over hand-crafted features, their accuracy has further improved year-over-year even exceeding human performance on this complex dataset \citeb{HePReLU2015,Russakovsky2014}. CNNs keep on expanding to more areas of computer vision and data analytics in general \citeb{Abu-El-Haija2016,Kaluarachchi2015,Zhang2016a,Park2016} and are moving towards analyzing video data for action recognition \citeb{Wang2015}, tracking \citeb{Chen2017a}, and improved object detection \citeb{Kang2017,Jie2018}. 

Unfortunately, the high accuracy of CNNs comes with a high computational cost, requiring powerful GPU servers to train these networks for several weeks using hundreds of gigabytes of labeled data. While this is a massive effort, it is a one-time endeavor and can be done offline for many applications. However, the inference of state-of-the-art CNNs also requires several billions of multiplications and additions to classify even low-resolution images by today's standards \citeb{Cavigelli2015}. 
While in some cases offloading to centralized compute centers with powerful GPU servers is also possible for inference after deployment, it is extremely costly in terms of compute infrastructure and energy. Furthermore, collecting large amounts of data at a central site raises privacy concerns and the required high-bandwidth communication channel causes additional reliability problems and potentially prohibitive cost of deployment and during operation \citeb{Ananthanarayanan2017}. \revision{For many applications, the introduced latency is prohibitive. }

The alternative, on-site near sensor embedded processing, largely solves the aforementioned issues by transmitting only the less sensitive, condensed information---potentially only security alerts in case of a smart surveillance camera---but imposes restrictions on available computation resources and power. 
These push the evaluation of such networks for real-time semantic segmentation or object detection out of reach of even the most powerful embedded platforms available today for high-resolution video data \citeb{Cavigelli2015}. However, exactly such systems are required for a wide range of applications limited in cost (CCTV/urban surveillance, perimeter surveillance, consumer behavior and highway monitoring) and latency (aerospace and UAV monitoring and defense, visual authentication) \citeb{Ananthanarayanan2017,Nguyen-Meidine2017}.

Large efforts have thus already been taken to develop optimized software implementations for heterogeneous platforms 
\citeb{Vasilache2018,Chetlur2014,Cavigelli2015,Lavin2015,Lavin2015a}, to design specialized hardware architectures \citeb{Cavigelli2016,Andri2016,Cavigelli2015a,Chen2016,Park2016,Farabet2011}, and to adapt the networks to avoid expensive arithmetic operations by reducing arithmetic precision, exploiting sparsity, and developing more compact DNNs \citeb{Rastegari2016,Zhang2016a}. \revision{However, they either 1) do not provide a strong enough performance boost, 2) are already at the theoretical limit of what can be achieved on a given platform, 3) are inflexible and not commercially available, or 4) incur a considerable accuracy loss. It is thus essential to extend the available options to efficiently perform inference on CNNs. }

\change{
In this paper, we propose a novel method performing \textbf{c}hange-\textbf{b}ased \textbf{infer}ence (hence named CBinfer) for convolutional neural networks on video data from a static camera with limited frame-to-frame changes. We extend our preliminary work in \citeb{Cavigelli2017}:

\begin{enumerate}
 \item Enhancements to the algorithm for improved compute time and ensuring a consistent input-output relation for each convolution layer.
 \item In-depth analysis how changes propagate through the CBinfer DNN. 
 \item Analysis of accuracy, compute time and energy efficiency for long frame sequences. 
 \item \revision{Additional evaluations for pose and object detection applications with much deeper networks and datasets without annotations.}
 \item Optimizations and evaluations on the more recent Nvidia Tegra~X2 platform\footnote{The Nvidia Tegra X2 is a system-on-chip available on an embedded board with an affordable power budget ($<15$\,W) for a stationary smart camera.}.
 \item Discussion and evaluation of the processing steps and how the chosen configuration provides the highest performance gain. 
\end{enumerate}
Overall the proposed method provides an average speed-up by a factor of 9.1--7.0 over an implementation relying on cuDNN and introducing only negligible accuracy loss. It thus significantly outperforms previous approaches exploiting frame-to-frame locality which all have measured performance gains in the range of a few ten percent while introducing accuracy losses of several percent (cf. \secref{sec:relWorkVideo}). Our method can be combined with most single-frame optimizations such as exploiting weight sparsity or the development of more compact DNN models. The code is available online at \url{https://github.com/lukasc-ch/CBinfer}.
}

\section{Related Work} \label{sec:relWork}
\change{In this section, we first describe existing optimized implementations for CNN inference and existing approximations trading accuracy for throughput. We then specifically survey related approaches exploiting the limited changes in video data to reduce the computational effort required to perform CNN inference. Finally, we discuss available datasets and CNNs with which we can evaluate our proposed algorithm.

Most per-frame optimization techniques can be combined with the method we propose herein. Existing approaches targeting video data have very limited gains and have not been specifically optimized for static camera frame sequences. }

\subsection{Optimized Embedded System Implementations} \label{sec:relWorkImpl}
The latest wave of interest in neural networks can be attributed to their sudden success driven by the availability of large datasets and the increasingly powerful computing platforms. One of the most economical and practicable solutions for training medium-sized CNNs is to use a workstation with GPUs. The available software frameworks to implement and train CNNs provide strong support for this kind of platform. 

The massive amounts of compute time spent training CNNs has spurred the development of highly optimized GPU implementations. First, most widely used frameworks relied on their own custom implementations which have all converged to methods relying on matrix-multiplications,  
leveraging the availability of highly optimized code in BLAS libraries and the fact that GPUs are capable of achieving a throughput within a few percent of their peak performance with this type of workload. Specialized libraries such as Nvidia's cuDNN and Nervana Systems' Neon provide some additional performance gains through assembly-level implementations \citeb{Lavin2015} and additional algorithmic improvements such as Winograd and FFT-based convolution \citeb{Lavin2015a}. A specific implementation for non-batched inference on an embedded platform building on a matrix multiplication is documented in \citeb{Cavigelli2015}, also showing that more than 90\% of time is spent computing convolutions.

\subsection{Approximations Trading Accuracy for Throughput} \label{sec:approxCNN}
\change{
DNNs commonly require a high computation effort in the order of 20\,GOp/frame for classification of a $224\times 224$ pixel image (1 multiply-add is counted as 2 operations) \citeb{Canziani2017}. Extracting features when working with high resolution images (e.g. for object detection or semantic segmentation) scales up the effort proportional to the number of pixels, quickly reaching few 100\,GOp/frame. 

Admitting limited accuracy losses in order to gain a higher throughput by approximating existing networks, inference algorithms, and arithmetic operations can help overcome the computational obstacles preventing widespread adoption of CNN-based algorithms on embedded and mobile platforms. 
Several such approaches are surveyed and compared in \citeb{Sandler2018,Iandola2017}. In this section, we will provide an overview of different options that can be exploited. 
}

% precision/arithmetic/approx. comp./ vector quant. / pruning
\change{
One such option is the reduction of the required arithmetic precision to evaluation NNs. Various methods from normal fixed-point analysis to retraining networks to adapt for quantized weights and activations exist. On some off-the-shelf software programmable platforms, 16-bit or 8-bit arithmetic operations can be vectorized to obtain a performance boost \citeb{Gysel2016a}. 
Extreme methods go as far as to enforce binary weights \citeb{AojunZhou2016,Andri2018}, and in some cases also binary activations \citeb{Rastegari2016}. 
}
This means that multiplications can be dropped entirely, and in case of binary activations even collapse some of the add/subtract operations into XNOR and bit count operations. Many networks can be quantized with 8\,bit without an increase in error rate, before there is a trade-off between precision and accuracy \citeb{Cavigelli2016,Hashemi2016}. Some methods try reducing the computational effort by pruning many very small weights to zero, making it possible to skip some operations \citeb{Li2016}, or even dynamically skip operations when the activations are zero \citeb{Aimar2017}. More sophisticated quantization schemes such as vector quantization exist and can further compress a trained CNN model, but they require specialized hardware to bring an improvement in energy efficiency \citeb{Han2016a,Aimar2017}. 

% network concepts (e.g. Faster R-CNN, YOLO, fully convolutional; tensorized convolutions; smaller networks)
Further research has focused on optimizing semantic segmentation and object detection algorithms to better reuse already computed features by eliminating any non-convolutional elements from the network \citeb{Redmon2016,Long2015} or introducing structured sparsity \citeb{Zhang2017}. Simplifying the operations in a network, such as low-rank approximations of 2D convolutions or by simply designing smaller networks with state-of-the-art methods have been evaluated in \citeb{Iandola2016,Canziani2017}. 

The method we propose in this paper does not supersede these methods, but can be combined with the aforementioned approximation methods to further improve throughput.

\subsection{Video-based Computation Reduction} \label{sec:relWorkVideo}
Obtaining per-frame features naturally seems like an easier task when these frames belong to a video sequence rather than a random collection of images. Limited movement of objects in a frame can be exploited in object tracking by working with a limited search window within the frame \citeb{Held2016}, not only reducing the problem size, but also simplifying the regression task---up until the tracked target is occluded by a large object.

\emph{Clockwork CNNs} \citeb{Shelhamer2016}
specifically target CNNs for semantic segmentation with a structure similar to \citeb{Long2015}. They have extended this work on fully convolutional networks, which presents a CNN with skip connections and deconvolution layers to refine the lower-resolution feature maps obtained deep within the network using the features extracted early in the network. They exploit that lower-resolution feature maps within the network are more stable over time than the full-resolution input. They thus propose to reevaluate the first few layers and the last layers affected through the skip connections more frequently than the coarser grained feature maps. This is a strong limitation on the set of CNNs this method can be applied to. They present evaluations based on a static as well as a dynamic, content-adaptive reevaluation schedule, showing that they can reduce the number of full-frame convolutions by about 40\% before the accuracy starts to drop on the Youtube-Objects dataset. 
However, this approach is limited to updating entire frames, whereas we exploit that often only small parts of the scene change and need to be reevaluated, which leads to larger savings. 

\change{
\emph{CNNCache} \citeb{Xu2017} describes a general approach pursuing a similar direction of work. They describe their method as a caching mechanism, where blocks of the image are matched to blocks in the previous frame, thereby fetching results of similar block from the cache instead of recomputing the results. Similarly to our work, this requires the selection of a threshold, and on top of that a block size and a cache depth in the form of an expiration time. The block matching allows to handle video data where the camera is not fully static, but it does not allow perspective changes. They have shown that their method achieves an average speed-up in the order of 20\% at a top-1 accuracy loss of 3.5\% performing image classification relative to the \emph{ncnn} framework's default implementation. 
The capability to recall convolution results even when the specific image tile has moved introduces a significant overhead comparing image tiles, thereby limiting the potential speed-up significantly. Further, this method requires a relatively high tolerance when comparing image tiles to be able to find matches, thereby introducing significant accuracy losses. 

\emph{DeepMon} \citeb{Huynh2017} proposes another method combining convolution layer decomposition, half-precision computation, and convolutional layer caching. Similarly to CNNCache, they divide the input to each convolutional layer into blocks and reuse the result when a block matches to the one in the previous frame. To reduce overhead, they do not directly compare the blocks, but instead extract histogram-based features.
They apply their technique only to the first few layers, because in later layers the caching overhead exceeds the compute latency savings. 
They show a speed-up attributable to caching of 18\% for object detection and 36\% for image classification at an accuracy loss in the order of 3.8\% to 6.2\%. 
While their histogram-based comparison method for the image tiles reduces overhead, it still remains significant and the introduced accuracy loss increases further. 

\emph{Sigma-Delta Quantized Networks} \citeb{Connor2017} is the most similar method to ours. They combine quantizing the network and decomposing the input to each convolution layer with the difference of the current frame's values to the previous frame's values and accumulate the result over time. They show a $4-10\times$ reduction in the number of operations in total, of which $2-3\times$ can be attributed to the temporal differences aspect of their method at an accuracy drop. However, whether this reduction in number of multiply-add operations can be put into performance gains after all the introduced overhead remains an open question. 
}

\subsection{Suitable Datasets and Neural Networks}
\label{sec:dataset}
\begin{table}[b]
    \centering
    \caption{Semantic Segmentation CNN Used For Evaluations.}
    \label{tbl:convnet}
    \small
    \addtolength{\tabcolsep}{-4pt} 
    \change{
    \begin{tabular}{@{}clccrr@{}}
        \toprule
          & Type & Outp. Res. & Feat. Maps & CT [ms] & rel. CT \\ 
         \midrule
         L1 & conv $7\times 7$ & $541\times 871$ & $3\rightarrow 16$ & 72.9 & 13.6\% \\ 
         L2 & act., pool $2\times 2$ & $271\times 436$ & $16\rightarrow 16$ & 10.7 & 2.0\% \\
         L3 & conv $7\times 7$ & $271\times 436$ & $16\rightarrow 64$ & 116.2 & 21.7\% \\ 
         L4 & act., pool $2\times 2$ & $136\times 218$ & $64\rightarrow 64$ & 10.2 & 1.9\% \\
         L5 & conv, act. $7\times 7$ & $136\times 218$ & $64\rightarrow 256$ & 309.4 & 57.8\% \\ 
         L6 & conv, act. $1\times 1$ & $136\times 218$ & $256\rightarrow 64$ & 14.4 & 2.7\% \\ 
         L7 & conv $1\times 1$ & $136\times 218$ & $64\rightarrow 8$ & 1.6 & 0.3\% \\ 
         \midrule
         & Total & & & 535.4 & \\
         \bottomrule 
    \end{tabular}
    }
    \addtolength{\tabcolsep}{+4pt}  
\end{table}

\change{
We show the applicability of the concept to various applications, namely by evaluating the proposed method for semantic segmentation and pose detection. These are both often applied to high-resolution images and video streams with high frame rates above 10\,frame/s for meaningful applications. }

\begin{figure}
	\centering
	\includegraphics[width=\linewidth]{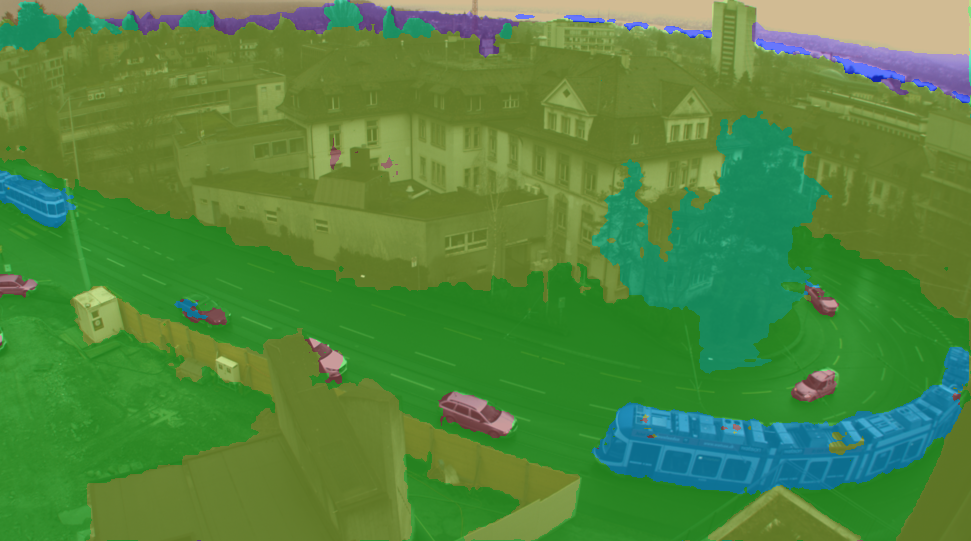}
	\caption{Example output of the scene labeling network of \citeb{Cavigelli2016a} on which we evaluate our algorithm.}
	\label{fig:labelling}
\end{figure}

\begin{figure}
  \centering
  \includegraphics[width=\linewidth]{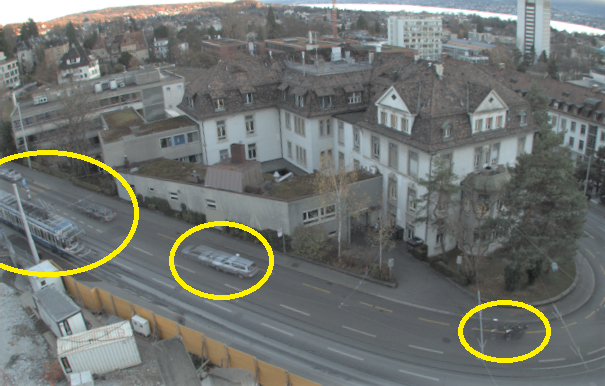}
  \caption{A sample video sequence from the dataset of \citeb{Cavigelli2016a} showing the frame-by-frame changes by overlaying a sequence of length 3. Moving objects are only a small part of the overall scene and affect only a small share of the pixels.}
  \label{fig:gloria_seq}
\end{figure}

\change{We are specifically interested in video sequences obtained from a static camera. While some such datasets exist (e.g. person or vehicle detection or re-identification), most of them are limited to extremely few (1-3) classes and rarely target semantic segmentation. }
However, for first application scenario---semantic segmentation---the dataset\footnote{Available online at \url{https://doi.org/10.3929/ethz-b-000276417}} used in \citeb{Cavigelli2016a} provides ground truth labels for 10-class semantic segmentation from an urban street surveillance perspective, and while they work with individual images, several surrounding unlabeled frames and a trained convolutional network are available. An example image labeled with the provided CNN is shown in \figref{fig:labelling}, and a sample sequence of 3 images is visualized in \figref{fig:gloria_seq}.

\begin{figure}
  \change{
  \centering
  \includegraphics[width=\linewidth]{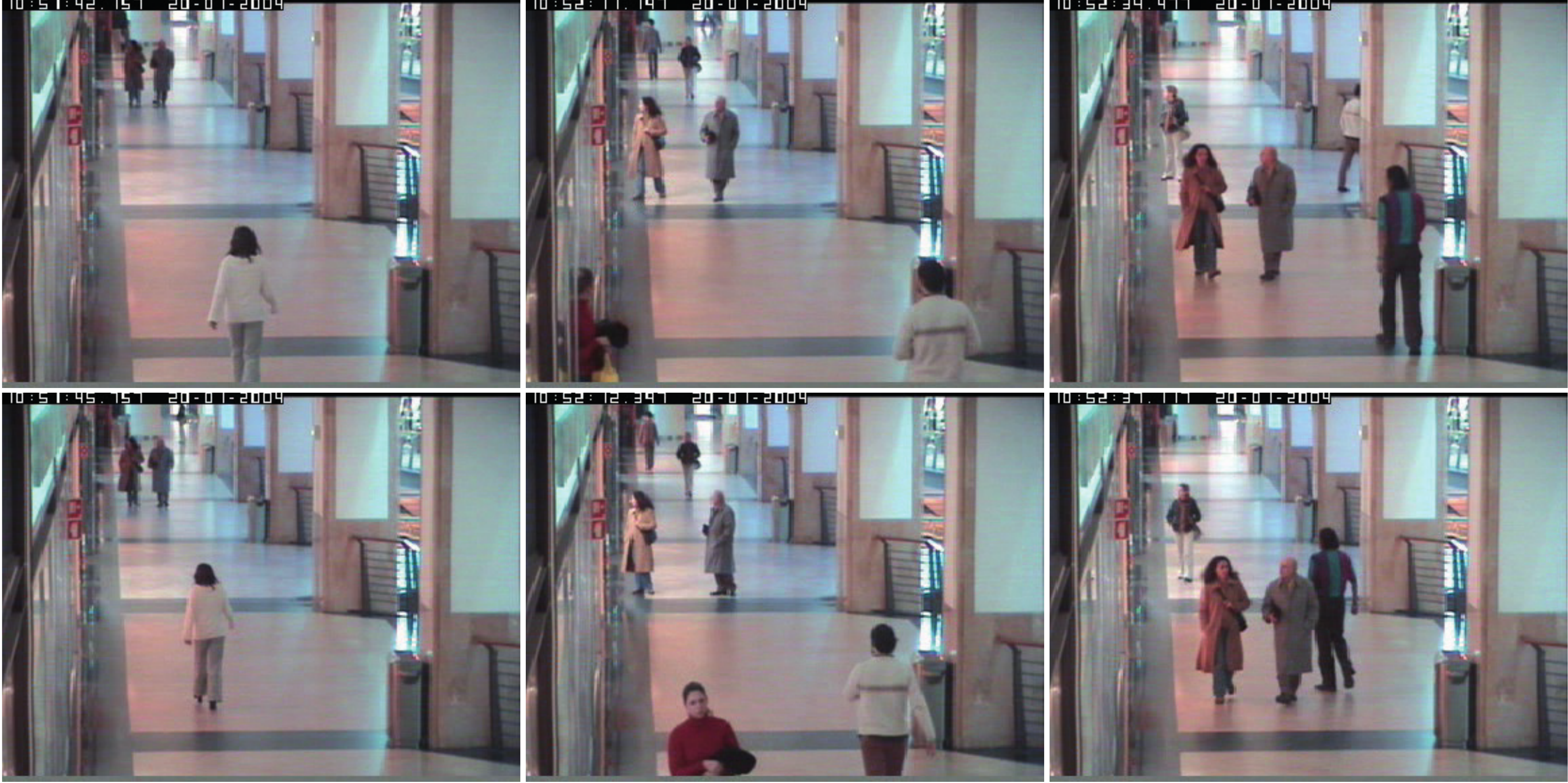}
  \caption{Sample frames from sequences of the CAVIAR dataset on which we perform evaluations for pose detection.}
  \label{fig:caviar}
  }
\end{figure}

\change{
For the second application---pose detection---several datasets to detect joints and limbs exist in the form of annotated images or a moving camera frame sequences, but none with a static camera. To overcome this and to show the feasibility of applying CBinfer without annotated data, we use unlabeled frame sequences from the CAVIAR dataset\footnote{\change{Available at \url{http://homepages.inf.ed.ac.uk/rbf/CAVIAR/}, collected through the EC Funded CAVIAR project/IST 2001 37540.}} and take the pretrained network to generate the reference output. The dataset contains scenes recorded using surveillance cameras with wide-angle lenses and captures the interaction of few people. It has a resolution of $384\times 288$ pixel and a frame rate of 25\,frame/s. A few sample frames are shown in \figref{fig:caviar}. 
}

\revision{For object detection---our third application scenario---we use video sequences of traffic surveillance cameras. Object detection is performed using YOLOv3 \cite{Redmon2018} trained on the MS COCO dataset \cite{Lin2014}. Since there is no ground truth available for the sequences, we generate our reference output by applying the original YOLOv3 network to each frame. }
\begin{figure}
  \revision{
  \centering
  \includegraphics[width=\linewidth]{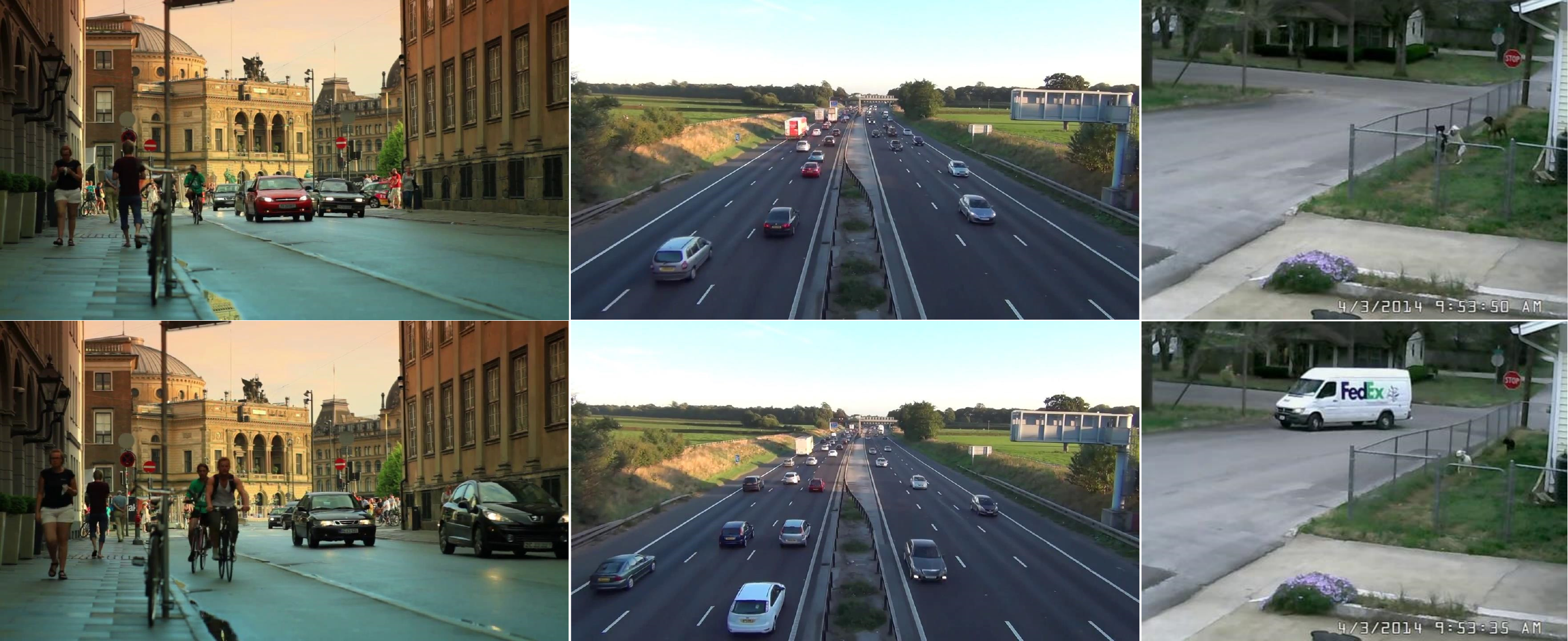}
  \caption{Sample frames from the video sequences on which we perform evaluations for object detection.}
  \label{fig:objdetFrames}
  }
\end{figure}

\section{Methodology} \label{sec:method}
%Differently from to previous work looking at reevaluating entire frames, we exploit the limited number of pixels changing frame-to-frame to increase the throughput without loss in classification accuracy. 
The most straight-forward pixel-level approach is to detect changing pixels on the input frame based on a threshold on the difference to the previous frame, and then update all the pixels affected by them. This increases the number of pixels to be updated layer-after-layer due to the convolution operations. Thus for e.g. a $7\times 7$ convolution, a one-pixel change triggers an update of 49 pixels in the next layer and 169 pixels after another $7\times 7$ convolution. Strided operations (often used with pooling layers) reduce this effect, but do not prevent it. This issue might seem prohibitive for multi-layer CNNs, particularly when considering that individual pixels might keep exceeding the threshold due to noise. 

However, the change is not only spatially local at the input, but also at the output. Furthermore, noise-like changes will likely not have strong impacts on feature maps deeper within the network. We thus propose to perform the change-detection not only at the input, but before each convolution layer---relative to its previous input---and to compute an updated value only for the affected output pixels. This can be done without modifications to the training of the CNN, can be applied to existing pre-trained networks, and is not specific to the CNN on which we evaluate the proposed algorithm. 

We propose to replace all spatial convolution layers (conv layers) with \emph{change-based} spatial convolution layers (CBconv layers). This means adapting the widely used, simple and well-performing matrix-generation and matrix-multiplication sequence of operations \citeb{Jia2013,Cavigelli2015}. The convolution layer computes

\begin{small}
\revision{
\begin{align}
    y_o(j,i) = b_o + \sum_{c\in\mathcal{C}_{in}}\sum_{(\Delta j,\Delta i)\in\mathcal{S}_k} k_{o,c}(\Delta j,\Delta i) x_c(j-\Delta j,i-\Delta i),
\end{align}
}
\end{small}
where $o$ indexes the output channels $\mathcal{C}_{out}$ and $c$ indexes the input channels $\mathcal{C}_{in}$. The pixel is identified by the tuple $(j,i)$ and $\mathcal{S}_k$ denotes the support of the filters kernels $k$. 
This can be computed by performing a matrix multiplication
\revision{
\begin{align}
	\mx{Y} = \mx{K} \mx{X}&, 
	\quad \mx{Y}\in\mathbb{R}^{|\mathcal{C}_{out}| \times h_o \cdot w_o}, \nonumber\\
	\quad \mx{K}\in\mathbb{R}^{|\mathcal{C}_{out}| \times |\mathcal{C}_{in}| \cdot h_k \cdot w_k}&, 
	\quad \mx{X}\in\mathbb{R}^{|\mathcal{C}_{in}| \cdot h_k \cdot w_k \times h_o \cdot w_o}.\label{eq:matMult}
\end{align}
}
The image matrix $\mx{X}$ is constructed as $X((k h_k + j)w_k + i, y_o w_o + x_o) = x(k, j+y_o, i+x_o)$ with $k=1,\dots,|\mathcal{C}_{in}|$, $j=1,\dots,h_k$, $i=1,\dots,w_k$ and $y_o=1,\dots,h_o$, $x_o=1,\dots,w_o$. 
The filter matrix $\mx{K}$ is given by $K(o, (c h_k + j) w_k + i) = k(o,c,j,i)$ for $o=1,\dots,|\mathcal{C}_{out}|$, $c=1,\dots,|\mathcal{C}_{in}|$, $j=1,\dots,h_k$ and $i=1,\dots,w_k$. 
The result matrix is stored as $Y(o, y_o w_o + x_o) = y(o, y_o, x_o)$. Zero-padding can be applied during the construction of the $\mx{X}$ matrix and an efficient strided convolution can be computed by dropping the unused rows. 

We replace this matrix multiplication by the following sequence of processing steps, thereby drastically reducing the size of the matrix used in the main computation step. 

\subsection{Processing Steps} \label{sec:procSteps}
\begin{figure*}
  \centering
  \definecolor{babyblueeyes}{rgb}{0.63, 0.79, 0.95}
  \includegraphics[width=\linewidth]{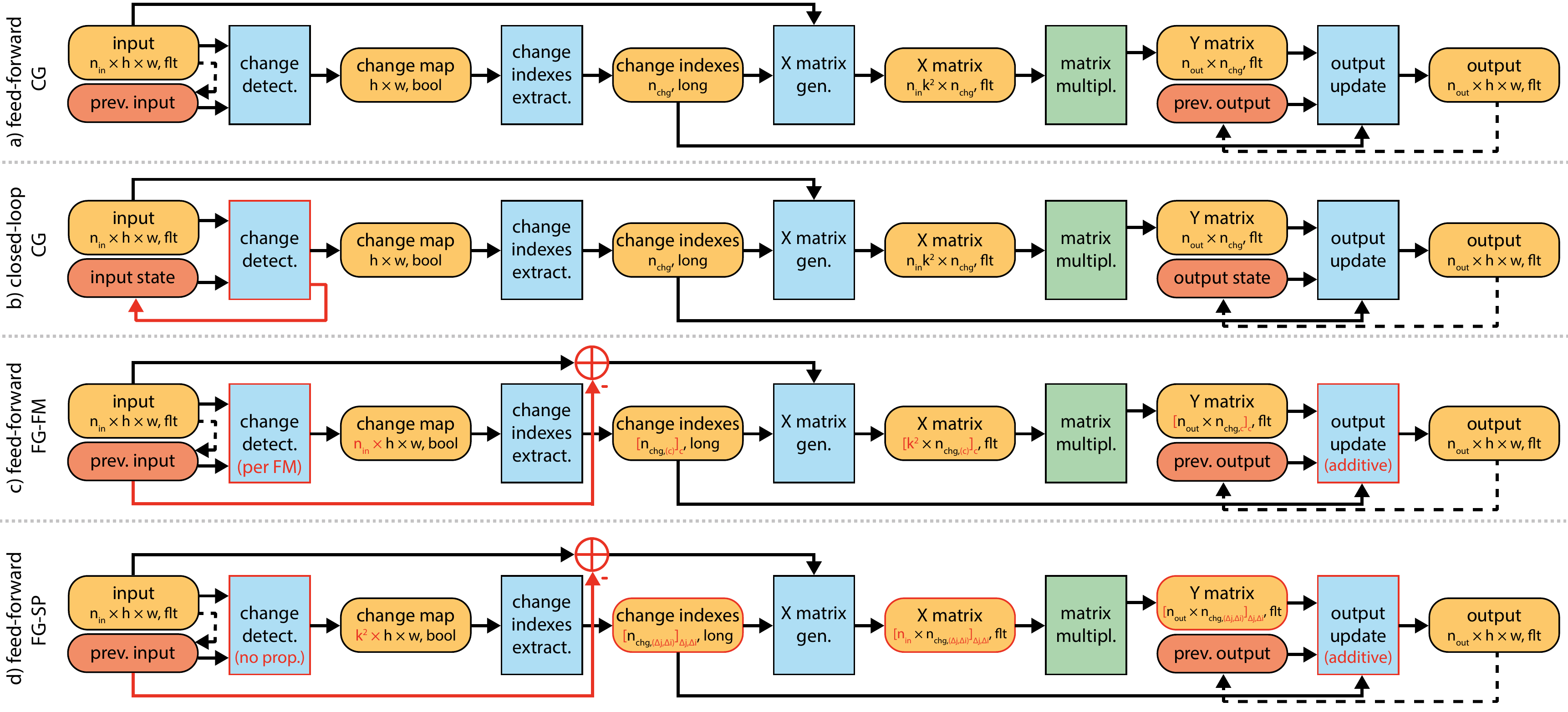}%4
  \revision{
  \caption{Processing flow and intermediate data tensors of CBinfer. Color code: \textcolor{babyblueeyes}{custom processing kernels}, \textcolor{green!60!black!60!}{cuBLAS kernel}, \textcolor{yellow!20!orange!80!}{variables sharable among layers}, and \textcolor{red!40!orange!60!}{variables to be stored per-layer}. Size and data type of intermediate results are indicated below the variable name. Coarse-grained feed-forward CBinfer (a) is introduced in \secref{sec:procSteps}, the closed-loop formulation (b) is described in \secref{sec:closedLoop}, and the fine-grained extensions to the algorithm (c,d) are formulated in \secref{sec:fineGrained}.}
  \label{fig:algostruct}
  }
\end{figure*}
We modify the standard approach and use a sequence of processing steps (cf. \figref{fig:algostruct}, top/\emph{feed-forward}): change detection, change indexes extraction, matrix generation, matrix multiplication, and output update. In the following, we will explain the individual steps. 
\paragraph{Change Detection}
In this step, changed pixels are detected. We define a changed pixel as one where the absolute difference of the current to the previous input of any feature map/channel exceeds some threshold $\tau$, i.e.
\begin{align}
    m(j,i) = \bigvee_{c\in\mathcal{C}_I}\left(\abs{x^{(t)}(c,j,i)-x^{(t-1)}(c,j,i)}>\tau\right). \label{eq:changeDet1}
\end{align}
The computation effort of this step is crucial, since it is executed independently of whether any pixel changed. Each of these changes affects a region equal to the filter size, and these output pixels are marked for updating: 
\begin{align}
    \widetilde{m}(j,i) = \bigvee_{(\Delta j,\Delta i)\in\mathcal{S}_k} m(j+\Delta j,i+\Delta i), \label{eq:changeDet2}
\end{align}
where $\mathcal{S}_k$ is the filter kernel support, e.g. $\mathcal{S}_k=\{-3,\dots,3\}^2$ for a $7\times 7$ filter. 
All of this is implemented on GPU by clearing the change map to all-zero and having one thread per pixel, which---if a change is detected---sets the pixels of the filter support neighborhood in the resulting \emph{change map}. 

\paragraph{Change Indexes Extraction}
In this step, we condense the change map $\widetilde{m}$ to 1) a list of pixel indexes where changes occurred and 2) count the number of changed pixels. This has been implemented by relying on the Thrust\footnote{\url{https://thrust.github.io}} \texttt{copy\_if} function. 
The computed index list is later on needed to access only the needed pixels to assemble the matrix for the convolution. 

\paragraph{Matrix Generation \& Matrix Multiplication}
Matrix multiplications are used in many applications, and highly optimized implementations such as the GEMM (general matrix multiplication) function provided by the Nvidia cuBLAS library come within a few percent of the peak FLOPS a GPU is capable to provide. Matrix multiplication-based implementations of the convolution layer relying on GEMM are widely available and are highly efficient \citeb{Jin2014,Cavigelli2015} as described above. 
The $\mx{X}$ matrix in \eqref{eq:matMult} is not generated full-sized, but instead only those columns corresponding to the relevant output pixels are assembled, resulting in a reduced width equal to the number of output pixels affected by the changes in the input image. 
\revisionB{The columns to be generated are selected using the \emph{change indexes} (cf. \figref{fig:algostruct}) and are constructed following the procedure described in the previous section. This is implemented with independent threads for each pixel and spatial filter position, where each of them copies all the feature map values at the position.}

The $\mx{K}$ matrix is made up of the filters trained using normal convolution layers and keeps the same dimensions, so the computation effort in this step is proportional to the number of changed pixels and the matrix multiplication is in the worst case only as time consuming as the full-frame convolution. 

\paragraph{Output Updating}
We use the previously stored results and the newly computed output values along with the change indexes list to provide the updated output feature maps. To maximize throughput, we also include the ReLU activation of the affected pixels in this step, reducing the compute time by 1) not writing the value to memory and immediately reading them again---an independent ReLU layer is strongly memory bandwidth limited, and 2) only applying the ReLU operation to changed pixels.

\subsection{Memory Requirements}
The memory requirements of DNN frameworks are known to be very high, up to the point where it becomes a limiting factor for increasing the mini-batch size during learning and thus reducing the throughput when parallelizing across multiple GPUs. These requirements are very different when looking at embedded inference-only systems:
\change{ 
\begin{enumerate}
    \item \revision{Inference is typically done on single frames. Creating mini-batches would introduce often unacceptable latency while only providing a few percent of additional performance \citeb{Cavigelli2015}. }
    \item During training, the input of each layer has to be stored in order to be able to compute the gradients. This is not required during inference. 
    \item Batch normalization layers, Dropout layer, etc. (if present) are considered independent layers during training. They can be absorbed into the convolution layer for inference. 
\end{enumerate}
}
To obtain a baseline memory requirement, we compute the required memory of common DNN frameworks performing convolutions using matrix multiplication with a batch size of 1. We assume an optimized network minimizing the number of layers, e.g. by absorbing batch normalization layers into the convolution layers or using in-place activation layers. This way 30M values need to be stored for the intermediate results, 264M values for the $\mx{X}$ matrix, and 873k values for the parameters. This can further be optimized by sharing  $\mx{X}$ among all convolution layers and by keeping only memory allocated to storing only the output of two layers and switching back-and-forth between them, layer-by-layer. This reduces the memory footprint to 9M, 93M, and 872k values, and a total of 103M values for our baseline. 

Applying our algorithm requires a little more memory, because we need to store additional intermediate results (cf. \figref{fig:algostruct}) such as the change matrix, the changed indexes list, and the $\mx{Y}$ matrix, which can all again be shared between the layers. We also need to store the previous output to use it as a basis for the updated output and to use it as the previous input of the subsequent layer. For our sample network, this required another $\sim$\,60M values to a total of 163M values (+\calc{0}{100/103*(103+60)-100}\%, total size $\sim$\,650\,MB)---an acceptable increase and not a limitation, considering that modern graphics cards typically come with around 12\,GB memory and even GPU-accelerated embedded platforms such as the Nvidia Jetson~TX2 module provide 8\,GB of memory. 

\change{
\subsection{Closed-Loop Formulation} \label{sec:closedLoop}
In \figref{fig:algostruct}a and \secref{sec:procSteps} we describe the processing steps for a \emph{feed-forward} implementation of CBinfer. 
However, note that this structure allows gradually changing inputs (e.g. two images are morphed over several frames with increments below the change detection threshold) to never trigger any update within the network and thus keep a stale result. 
In an outdoors surveillance setting, the effects could be even worse: consider a static scenery with a sunset and thus gradually changing brightness without triggering any update operation. Now a moving object passes, leaving a dark trace behind which has been updated under the changed lighting conditions. 

%closed loop structure
To overcome such issues, we are proposing a closed-loop version of CBinfer as shown in \figref{fig:algostruct}, bottom/\allowbreak\emph{closed-loop}. Rather than storing the previous input, we now have an \emph{input state}, which is updated only for those pixels which have triggered a change. This can be done directly in the change detection phase. 
This way, the previous output is consistently the convolution result of the input state and ensured not to drift far away from ideal result. 

%memory copying advantage
Since the previous input had to be stored before as well, this does not introduce any memory overhead. Moreover, in many cases it can even decrease compute time since only the few values where changes occurred have to be copied over from the \emph{input} to the \emph{input state}. For the feed-forward CBinfer the entire input tensor would have to be copied\footnote{\change{Note that one such tensor always has to be copied when applying CBinfer. Consider two CBinfer layers after each other. During the \emph{update output} step of the first CBinfer layer, we copy the newly computed values into the \emph{previous output} tensor and feed it to the next CBinfer layer as the \emph{input}. If we would not copy the data from input to \emph{previous input} here and instead just keep the memory address of the previous frame's input, it will be at the same location where the output of the first CBinfer layer's result will be stored when processing the next frame, thus directly modifying the \emph{previous input} variable and thereby introducing incorrect behavior (i.e. there are never any changes, since ultimately the \emph{input} and \emph{previous input} would point to the same memory location)}}. 
}

\subsection{\change{Fine-Grained Change-based Inference}} \label{sec:fineGrained}
\revision	{In the proposed scheme, every output value affected by any change at the input is recomputed.
As the convolution operation is linear, updates based on the difference to the previous frame can be computed to reduce the number of multiplications and additions in two ways:
\begin{enumerate}
\item Fine-grained across feature maps (FG-FM): Only some of the input feature maps affecting a given output value might have changed. An incremental update of the affected feature maps based on the difference of the change in input values relative to the previous frame would be sufficient. This results in a 3D-tensor change map and a correspondingly long change indexes list, and---crucially---forces to decompose the large matrix multiplication into several smaller ones of which the results have to be added individually during the update output step (cf. \figref{fig:algostruct}c). \label{itm:fgFM}
\item Spatially fine-grained (FG-SP): Just because an output pixel is affected by an input pixel does not require that it is completely recomputed. With a $3\times 3$ filter, a single pixel marked as change would trigger the re-computation of 9 pixels. Also, here an incremental update based on differences is possible (cf. \figref{fig:algostruct}d). \label{itm:fgSP}
\end{enumerate}
However, there are some drawbacks and limitations: 
\begin{itemize}
\item For both approaches the structure of the core computation effort is less regular and cannot be written as a dense matrix multiplication. 
\item The compute effort of the \emph{change indexes extraction} scales linearly with the number of values that have to be checked for changes. In case of (\ref{itm:fgFM}), the effort in this step is thus scale up by a factor of the number of input feature maps. 
\item The potential gains in case of (\ref{itm:fgSP}) are limited. Changing pixels are typically clustered together and all that is being saved is a small halo on the change map around the changes. This can in most cases be expected to be in the range of a few percent. 
\end{itemize}
}

\change{
\subsection{Propagating Changes \& Pooling} \label{sec:changeProp}
\begin{figure}
  \centering
  \includegraphics[width=\linewidth]{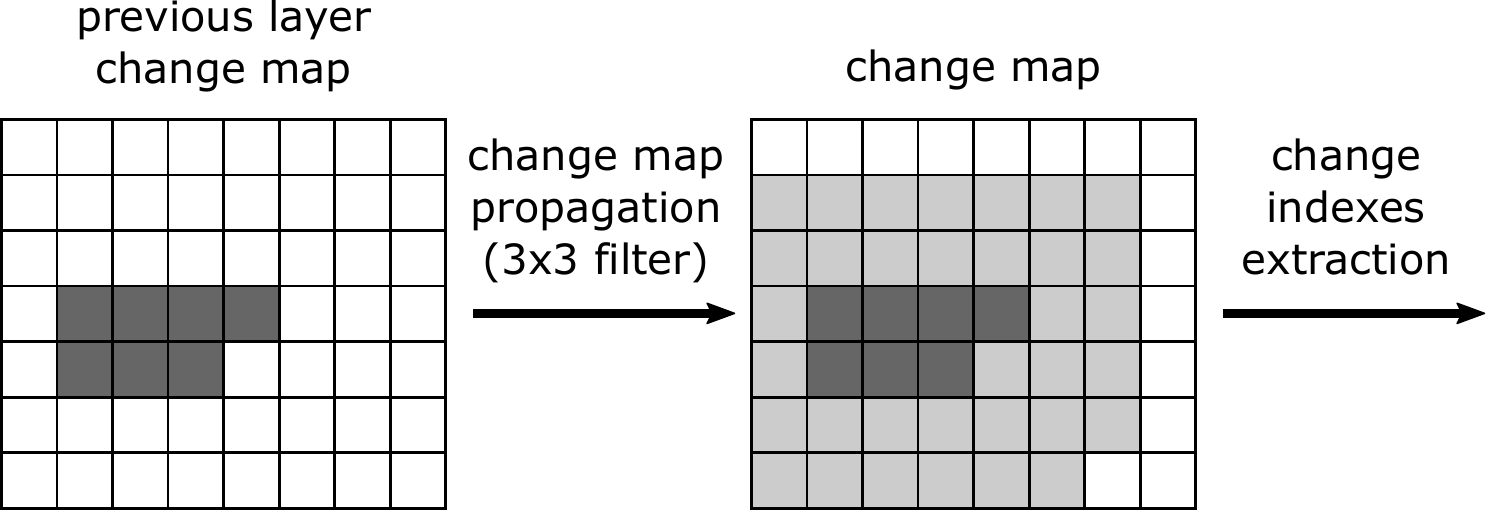}
  \revision{
  \caption{Worst-case propagation of change map when skipping the change detection step for a $3\times 3$ convolution. }
  \label{fig:changeMapProp}
  }
\end{figure}
\revision{
Change detection and change indexes extraction can contribute up to half of the compute time (cf. \secref{sec:compTimeBreakdown}). 
In some cases, it is thus worth considering skipping these steps:  
\begin{enumerate}
 \item If the previous layer was a CBconv layer as well, we can skip the change detection step and instead start from the previous layer's change map and apply change propagation to it (cf. Eq.~\ref{eq:changeDet2} and \figref{fig:changeMapProp}). This change propagation can be computed much faster, because no iteration across all the feature maps is necessary.
 \item In the special case that the current layer has a $1\times 1$ filter size, the changes do not propagate. This implies that the change map is identical to the one of the previous layer, which allows to also skip the change indexes extraction and re-use the change indexes of the previous layer. 
\end{enumerate}
Avoiding change detection also implies saving the memory to store the previous input for that layer. Besides the aforementioned advantages, there are some potential drawbacks:
}
\change{
\begin{enumerate}
 \item In case of (1), only the change detection step can be avoided and replaced with a change propagation step, and the change indexes have to be extracted again. The changes spread out at every layer this is done, although the change detection threshold might not have been exceeded everywhere and some of the changes could have been discarded. 

 \item For (2), there is no propagation of changes and both, change detection and change indexes extraction, can be skipped. So, the only drawback is that a few changes might be updated although they would be discarded if the input would be checked against the current layer's threshold.
\end{enumerate}
Besides for accelerating convolution layers, the above is also interesting for pooling layers which can also be implemented using a change-based approach. Since they typically follow a convolution layer, case (2) can be applied and the change-based update introduces no significant overhead but saves compute time---mostly by reducing memory bandwidth as pooling layers are memory-bound operations. 
}
}

\subsection{Threshold Selection} \label{sec:thSelectionProc}
\change{
The proposed algorithm adds one parameter to each convolution layer, the change detection threshold. \revision{It is fixed offline based on sample video sequences which are passed through the trained network. Other than through the selected values for the thresholds, this selection process does not affect the performance of the system. } A threshold of zero yields identical results to the non-change-based implementation, which has been used for functional verification. 

%definition/specification
For our evaluations, we perform an automated threshold selection process. First, all convolution layers are converted to change-based convolutions, and batch normalization and ReLU layers are absorbed into the CBinfer layers wherever possible. 
We define and choose:
\begin{enumerate}
 \item a performance metric such as pixel-wise classification accuracy, intersect-over-union (IoU), mean average precision (mAP)---possibly, the loss function of the network, 
 \item a set of frame sequences to evaluate the network, where the last frame is ideally annotated. An obvious alternative in case of a lack of frame sequences with annotated last frame is the generation the comparison of the change-based network model's output to the output of the original model using an appropriate metric, and
 \item an initial threshold, a factor determining the rate with which we adjust the threshold, and a maximum acceptable increment in quality loss per layer. 
\end{enumerate}

%procedure
We then set all thresholds to zero and start to iteratively step from the first to the last layer of the network. For each layer, we set an initial threshold value and evaluate the model with the aforementioned metric and dataset. We increment the threshold by a fixed factor (e.g. 1.1), re-evaluate, and repeat until the quality loss introduced by the current layer (with respect to a zero threshold) exceeds the maximum acceptable limit and then take the previous threshold value.

%non-linear network
In case of a DNN with (re-)convergent paths, we perform the threshold selection on these paths independently while setting the thresholds for the other paths to zero. 

%accuracy drop threshold per layer
The maximum acceptable quality loss can be set equally for all layers of the network. We focus on low accuracy loss configurations, and thus we are trying to select the threshold values such that they are right at the point where implementation losses are starting to occur. Nevertheless, we have observed best results by splitting the overall acceptable loss unevenly, allowing the first layer to introduce most of the loss. 
}

\section{Results \& Discussion} \label{sec:results}
\change{
In this section, we will first present the evaluation environment and analyze the baseline compute time breakdown. We then analyze the threshold selection, the effect on accuracy and achievable throughput. We then perform a more in-depth analysis of the throughput to verify the quality of the GPU implementation and investigate how the changes propagate in the network. We then establish why more fine-grained change detection does not pay off and how implementation loss and performance gains behave on longer sequences. 
}

\subsection{Evaluation Environment}
% dataset, network
\begin{figure}
  \centering
  \includegraphics[width=0.95\linewidth]{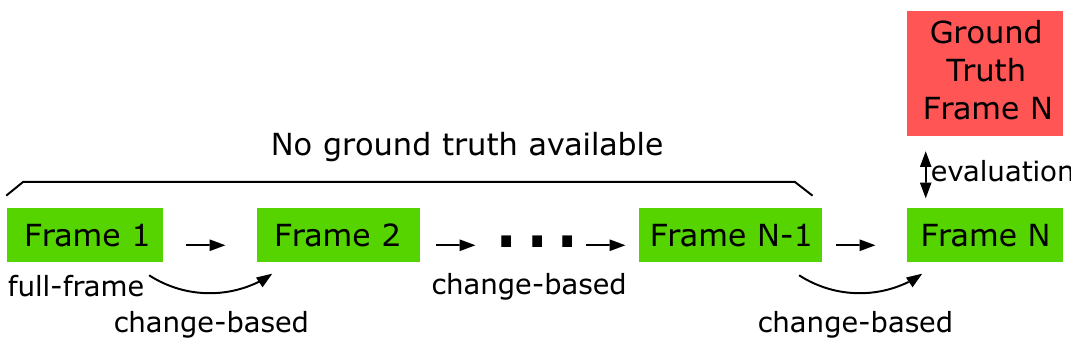}
  \caption{Scheme of the image sequence used for the evaluations.}
  \label{fig:seqSchema}
\end{figure}
\change{
We evaluate our method for two application scenarios: semantic segmentation and pose detection. 
For the first, we perform our evaluations on the urban surveillance dataset described in \secref{sec:dataset} and \citeb{Cavigelli2016a} and using the corresponding scene labeling CNN, not using the multispectral imaging data. The dataset provides 51 training images and 6 validation images with $776\times 1040$ pixel with the corresponding ground-truth scene labeling, classifying each pixel into one of the following 8 classes: building, road, tree, sky, tram, car/truck, water, distant background. 
For the validation set, the labeled images are part of short video sequences with 5 additional frames available before the frame for which the ground truth labeling is available. 
A trained network on this data is described in \citeb{Cavigelli2016a} and its parameters are reused unaltered for our evaluations. 
The procedure with which we perform our evaluations is visualized in \figref{fig:seqSchema}.

For the pose detection application, we use frame sequences from the CAVIAR dataset without ground truth annotations and the trained body estimation network of OpenPose \citeb{Cao2017} with $T=2$ stages. The frames are re-sampled to $368\times 490$ pixel as in the original OpenPose implementation to enable a meaningful comparison. The frame sequences are subsampled in time by a factor of 6 to arrive at a frame rate of around 4\,frame/s. In this setting, we measure the accuracy loss in terms of mean-squared error (MSE) relative to the output of the non-change-based network. We have found a MSE of $2\cdot 10^{-4}$ on the network's output to be sufficient for the pose detection to work reliably. With this dataset we run change-based inference for 9 frames before the accuracy and throughput measurements are performed on Frame 10 to avoid any start-up transients. As we will show later in \figref{fig:timeTrace}, these transients are very short and the error does not accumulate over time. 
}

\revision{For our experiment on object detection, we use the YOLOv3 network trained on the MS COCO dataset with 80 classes of everyday objects. The input image is rescaled, such that its smaller dimension corresponds to 416 pixels. The input sequences for our evaluations are described in \secref{sec:dataset}. Similar to pose detection we do not have ground-truth data, instead we generate our target output using non-change-based YOLOv3. For measuring the quality of the output feature maps, the MSE is not a suitable measure given that e.g. the outputs for the classification of the recognized object is scaled differently than the objectness score or the bounding box size. We have experimentally identified the objectness score to be the most sensitive output to potential artifacts of applying CBinfer and are thus measuring the accuracy loss due to change-based inference using the MSE on the objectness score. 
}

% framework, CUDA, machine
\change{We have implemented the proposed algorithm in the PyTorch framework using custom CUDA kernels, including functions to aid in converting DNNs to CBinfer (automatic conversion and threshold selection).
We have evaluated the performance on a Jetson TX2 board. 
Our performance baseline is the PyTorch implementation using Nvidia's cuDNN backend.} It includes optimizations such as the Winograd algorithm and FFT-based convolutions mentioned in \secref{sec:relWorkImpl}. Our evaluations were conducted using half-precision floating point numbers which have no negative impact on accuracy for both DNNs. 

\subsection{Baseline Throughput and Computation Breakdown} \label{sec:baselineComputeBreakdown}
\begin{table}
    \centering
    \caption{Performance Baseline Compute Time Breakdown}
    \label{tbl:baselineComputeTime}
    \begin{tabular}{c|rrr|rr}
        \toprule
         Layer & Conv. & Activ. & Pooling & total & share \\ \midrule
         1 & \ms{72.9} & \ms{7.4} & \ms{3.3} & \ms{83.6} & 15.6\% \\
         2 & \ms{116.2} & \ms{6.9} & \ms{3.3} & \ms{126.4} & 23.6\% \\
         3 & \ms{302.8} & \ms{6.6} & --- & \ms{309.4} & 57.8\% \\
         4 & \ms{12.7} & \ms{1.7} & --- & \ms{14.4} & 2.7\% \\
         5 & \ms{1.6} & --- & --- & \ms{1.6} & 0.3\% \\ \bottomrule
    \end{tabular}
\end{table}
\change{
Before we discuss the performance of the proposed algorithm, we analyze the baseline throughput and compute time breakdown of the segmentation DNN in \tblref{tbl:baselineComputeTime}. Clearly, the convolution operations are dominant, taking up 94.5\% (\ms{506.2}) of the overall computation time (\ms{535}). This reaffirms the focus on the convolution layers and will later on show that after accelerating the convolution operation significantly, optimizations for activation and pooling become relevant.
}

\subsection{Threshold Selection} \label{sec:ThresholdSelResults}
\begin{figure*}
  \centering
    \begin{tikzpicture}
        \begin{groupplot}
            [
                group style = {
                    group size=3 by 1,
                    ylabels at=edge left,
                    yticklabels at=edge left,
                    horizontal sep=1mm,
                },
                ylabel={Error Incr. [\%]},
                ymin=-0.075, ymax=0.8,
                width=0.39\linewidth, 
                height=50mm,
                grid=major, 
                cycle list name=seqStyleList,
                %xlabel=Threshold
            ]
            \nextgroupplot[xlabel={Threshold for Layer 1 ($\tau_1$)},]
            \addplot+ [] table 
                [x=threshold, y=errorDiff, col sep=comma]
                {\plotdataDir{outp-eval2-mode_thr1-pictureset15-cb.csv}};
            \nextgroupplot[xlabel={Threshold for Layer 2 ($\tau_2$)},]
            \addplot+ [] table [x=threshold, y=errorDiff, col sep=comma]
                {\plotdataDir{outp-eval2-mode_thr2-pictureset15-cb.csv}};
            \nextgroupplot[xlabel={Threshold for Layer 3 ($\tau_3$)},]
            \addplot+ [] table [x=threshold, y=errorDiff, col sep=comma]
                {\plotdataDir{outp-eval2-mode_thr3-pictureset15-cb.csv}};
        \end{groupplot}
    \end{tikzpicture}
  \caption{Analysis of the increase in pixel classification error rate by selecting a certain change detect threshold. This analysis is conducted layer-by-layer, where the error increase of any layer includes the error introduced by the previous layers' threshold choice ($\tau_1=0.04, \tau_2=0.3, \tau_3=1.0$).}
  \label{fig:thSelection}
\end{figure*}
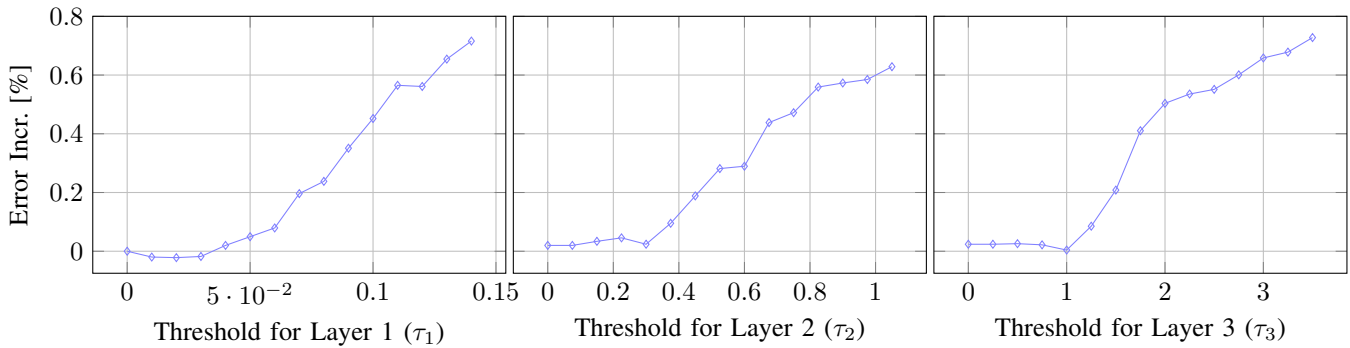
\change{
% threshold
Our algorithm introduces a threshold parameter for each layer, for which we outline the selection process in \secref{sec:thSelectionProc}. 
In \figref{fig:thSelection} we visualize the relation between accuracy and each layer's change detection threshold. We proceed similarly to our selection process, allowing an accuracy drop of 0.04\% per layer for the semantic segmentation network. Starting from all-zero thresholds ($\tau_i=0, i=1,\dots,3$), we sweep and select the optimal threshold parameter for each layer iteratively. The main purpose is to align the tipping points of the threshold-accuracy curve, such that not a single layer's threshold is limiting the overall accuracy. 
}

After the selection of the thresholds, we can scale them jointly to analyze the trade-off against the classification accuracy more concisely as can be observed in \figref{fig:allEvals} (left). The accuracy of the individual test sequences (different traces) clearly show a similar behavior with a plateau up to a clear point where there is a steep increase in error rate. \change{We repeated this analysis for the much deeper pose detection network (cf. \figref{fig:perfEvalsPoseDet}), showing similar behavior for the MSE with respect to the baseline DNN.}

\subsection{Throughput Evaluations}
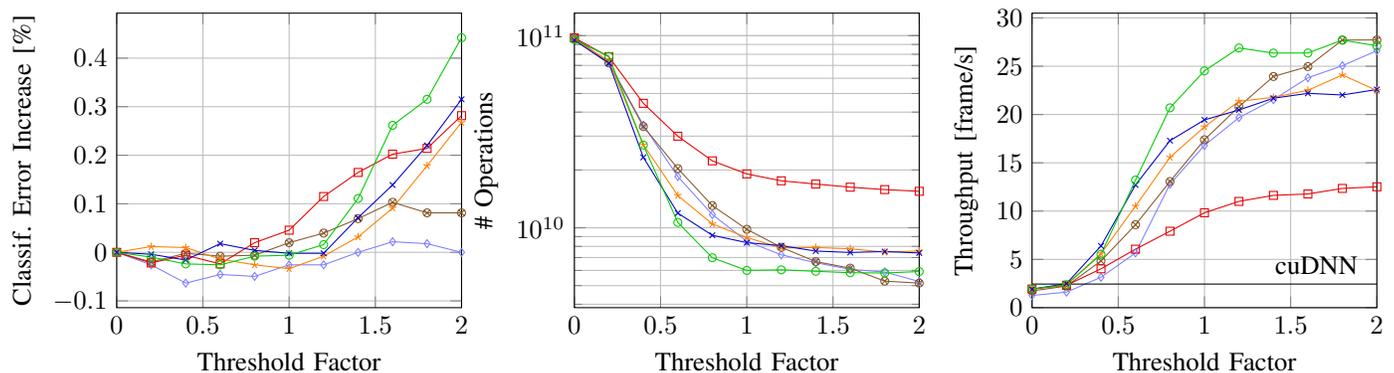
\begin{figure*}
	\change{
  \centering
    \begin{tikzpicture} 
        %read value for throughput baseline        
        \pgfplotstableread[col sep=comma]
            {\plotdataDirPfm{sceneLabeling-brienz2-eval01-single-baseline.csv}}\dataBaseline;
        \pgfplotstablegetelem{0}{throughput}\of\dataBaseline
        \pgfmathsetmacro{\baselineVal}{\pgfplotsretval}

        \pgfplotstableread[col sep=comma] {\plotdataDirPfm{sceneLabeling-brienz2-eval01-single-1607__11.csv}}{\dataTblA}
        \pgfplotstableread[col sep=comma] {\plotdataDirPfm{sceneLabeling-brienz2-eval01-single-1607__30.csv}}{\dataTblB}
        \pgfplotstableread[col sep=comma] {\plotdataDirPfm{sceneLabeling-brienz2-eval01-single-1611__12.csv}}{\dataTblC}
        \pgfplotstableread[col sep=comma] {\plotdataDirPfm{sceneLabeling-brienz2-eval01-single-1624__17.csv}}{\dataTblD}
        \pgfplotstableread[col sep=comma] {\plotdataDirPfm{sceneLabeling-brienz2-eval01-single-1624__48.csv}}{\dataTblE}
        \pgfplotstableread[col sep=comma] {\plotdataDirPfm{sceneLabeling-brienz2-eval01-single-1643__28.csv}}{\dataTblF}

        \begin{groupplot}
            [
                group style = {
                    group size=3 by 1,
                    horizontal sep=15mm,
                },
                width=0.34\linewidth, 
                height=55mm,
                cycle list name=seqStyleList, %seqColorList, 
                xmin=0, xmax=2, 
            ]
        %%%%% classification error
            \nextgroupplot[
                xlabel={Threshold Factor},
                ylabel={Classif. Error Increase [\%]}, 
                grid=major, 
                try min ticks=6, 
            ]
            \addplot+ table [x=th, y=loss] {\dataTblA};
            \addplot+ table [x=th, y=loss] {\dataTblB};
            \addplot+ table [x=th, y=loss] {\dataTblC};
            \addplot+ table [x=th, y=loss] {\dataTblD};
            \addplot+ table [x=th, y=loss] {\dataTblE};
            \addplot+ table [x=th, y=loss] {\dataTblF};
        %%%%% number of changes
            \nextgroupplot[
                xlabel={Threshold Factor},
                ylabel={\# Operations}, 
                grid=both, 
                try min ticks=6, 
                ymode=log,
            ]
            \addplot+ table [x=th, y expr=\thisrow{numGOps}*1e9] {\dataTblA};
            \addplot+ table [x=th, y expr=\thisrow{numGOps}*1e9] {\dataTblB};
            \addplot+ table [x=th, y expr=\thisrow{numGOps}*1e9] {\dataTblC};
            \addplot+ table [x=th, y expr=\thisrow{numGOps}*1e9] {\dataTblD};
            \addplot+ table [x=th, y expr=\thisrow{numGOps}*1e9] {\dataTblE};
            \addplot+ table [x=th, y expr=\thisrow{numGOps}*1e9] {\dataTblF};
        %%%%% throughput
            \nextgroupplot[
                xlabel={Threshold Factor},
                ylabel={Throughput [frame/s]}, 
                grid=major,
                try min ticks=6, 
                ymin=0,
            ]
            \addplot+ table [x=th, y=throughput] {\dataTblA};\label{trc:gloriastr_A}
            \addplot+ table [x=th, y=throughput] {\dataTblB};\label{trc:gloriastr_B}
            \addplot+ table [x=th, y=throughput] {\dataTblC};\label{trc:gloriastr_highact}
            \addplot+ table [x=th, y=throughput] {\dataTblD};\label{trc:gloriastr_C}
            \addplot+ table [x=th, y=throughput] {\dataTblE};\label{trc:gloriastr_D}
            \addplot+ table [x=th, y=throughput] {\dataTblF}; \label{trc:gloriastr_lowact}%\label{plt:thrghpt16}
            \draw[thin] (axis cs:\pgfkeysvalueof{/pgfplots/xmin},\baselineVal) -- (axis cs:\pgfkeysvalueof{/pgfplots/xmax},\baselineVal)
            node[above] at (axis cs:{\pgfkeysvalueof{/pgfplots/xmax}-0.35},\baselineVal) {cuDNN};
        \end{groupplot}
    \end{tikzpicture}
  \caption{Evaluation of the impact of jointly scaling the change detection thresholds on the classification error, the number of detected changed pixels (sum over all 3 layers), and the throughput. \revision{The various traces are different sequences of the Gloriastrasse segmentation dataset, where one sequence (\ref{trc:gloriastr_highact}) is a particularly active scene (road full of cars and trams), whereas in the sequence (\ref{trc:gloriastr_lowact}) shows a low-activity scene with a single car.} \revisionB{The traces correspond to Seq.~1643\_28~(\ref{trc:gloriastr_A}), Seq.~1607\_11~(\ref{trc:gloriastr_B}), Seq.~1611\_12~(\ref{trc:gloriastr_highact}), Seq.~1624\_48~(\ref{trc:gloriastr_C}), Seq.~1607\_30~(\ref{trc:gloriastr_D}), Seq.~1624\_17~(\ref{trc:gloriastr_lowact}) in the dataset.}}
  \label{fig:allEvals}
  }
\end{figure*}
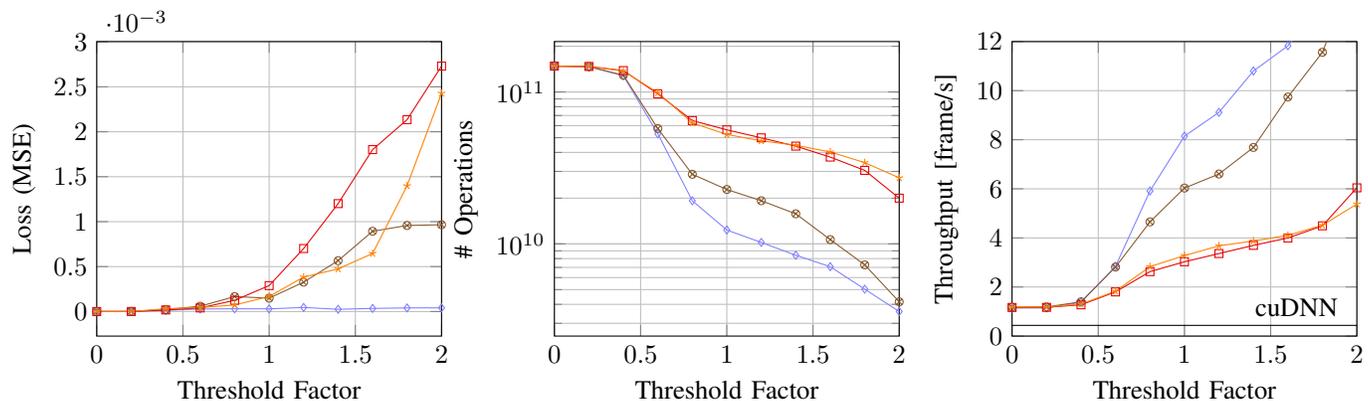
\begin{figure*}
	\change{
  \centering
    \begin{tikzpicture} 
        \pgfplotstableread[col sep=comma]
            {\plotdataDirPfm{poseDet-brienz2-eval01-half-baseline.csv}}\dataBaseline;
        \pgfplotstablegetelem{0}{throughput}\of\dataBaseline
        \pgfmathsetmacro{\baselineVal}{\pgfplotsretval}

        \pgfplotstableread[col sep=comma] {\plotdataDirPfm{poseDet-brienz2-eval01-half-caviar-walkByShop1cor-seq000-5xSubsample.csv}}{\dataTblA}
        \pgfplotstableread[col sep=comma] {\plotdataDirPfm{poseDet-brienz2-eval01-half-caviar-walkByShop1cor-seq001-5xSubsample.csv}}{\dataTblB}
        \pgfplotstableread[col sep=comma] {\plotdataDirPfm{poseDet-brienz2-eval01-half-caviar-walkByShop1cor-seq002-5xSubsample.csv}}{\dataTblC}
        \pgfplotstableread[col sep=comma] {\plotdataDirPfm{poseDet-brienz2-eval01-half-caviar-walkByShop1cor-seq003-5xSubsample.csv}}{\dataTblD}
        \def\poseDetPerfTables{
			\dataTblA, 
			\dataTblB, 
			\dataTblC, 
			\dataTblD}

        \begin{groupplot}
            [
                group style = {
                    group size=3 by 1,
                    horizontal sep=15mm,
                },
                width=0.34\linewidth, 
                height=55mm,
                cycle list name=seqStyleList,
                xmin=0, xmax=2
            ]
        %%%%% classification error
            \nextgroupplot[
                xlabel={Threshold Factor},
                ylabel={Loss (MSE)}, 
                grid=major, 
                try min ticks=6, 
            ]
            \addplot+ table [x=th, y=lossIncr] {\dataTblA};
            \addplot+ table [x=th, y=lossIncr] {\dataTblB};
            \addplot+ table [x=th, y=lossIncr] {\dataTblC};
            \addplot+ table [x=th, y=lossIncr] {\dataTblD};
        %%%%% number of changes
            \nextgroupplot[
                xlabel={Threshold Factor},
                ylabel={\# Operations}, 
                grid=both, 
                try min ticks=6, 
                ymode=log,
            ]
            \addplot+ table [x=th, y=numOps] {\dataTblA};
            \addplot+ table [x=th, y=numOps] {\dataTblB};
            \addplot+ table [x=th, y=numOps] {\dataTblC};
            \addplot+ table [x=th, y=numOps] {\dataTblD};
        %%%%% throughput
            \nextgroupplot[
                xlabel={Threshold Factor},
                ylabel={Throughput [frame/s]}, 
                grid=major,
                try min ticks=6, 
                ymin=0,ymax=12
            ]
            \addplot+ table [x=th, y=throughput] {\dataTblA};\label{trc:caviarA}
            \addplot+ table [x=th, y=throughput] {\dataTblB};\label{trc:caviarB}
            \addplot+ table [x=th, y=throughput] {\dataTblC};\label{trc:caviarC}
            \addplot+ table [x=th, y=throughput] {\dataTblD}; \label{trc:thrghpt16}
            \draw[thin] (axis cs:\pgfkeysvalueof{/pgfplots/xmin},\baselineVal) -- (axis cs:\pgfkeysvalueof{/pgfplots/xmax},\baselineVal)
            node[above] at (axis cs:{\pgfkeysvalueof{/pgfplots/xmax}-0.35},\baselineVal) {cuDNN};
        \end{groupplot}
    \end{tikzpicture}
  \caption{Evaluation of introduced loss (left), effective number of compute operations (center) and measured throughput (right) for several frame sequences running the pose detection network and varying the change detection threshold. \revisionB{The various traces correspond to different frame sequences of the CAVIAR dataset (cf. \figref{fig:caviar}). The traces correspond to Seq.~1~(\ref{trc:caviarA}), Seq.~2~(\ref{trc:caviarB}), Seq.~3~(\ref{trc:caviarC}), Seq.~4~(\ref{trc:thrghpt16}) in the dataset.}}
  \label{fig:perfEvalsPoseDet}
  }
\end{figure*}
The motivation for the proposed algorithm was to increase throughput by focusing only on the frame-to-frame changes. We show the performance gain in \figref{fig:allEvals} (right) with the indicated baseline analyzing the entire frame with the same network using cuDNN. In the extreme case of setting all thresholds to zero, the entire frame is updated, which results in a clear performance loss because of the change detection overhead as well as fewer optimization options such as less cache-friendly access patterns when generating the $\mx{X}$ matrix. \change{Nevertheless, few operations are skipped where the pixels did not change at all. }

When increasing the threshold factor, the average throughput increases rapidly to about \change{20\,frame/s}, where it starts saturating because the change detection step as well as other non-varying components like the pooling and pixel classification layers are becoming dominant and the number of detected changed pixels does not further decrease. We almost reach this plateau already for a threshold factor of 1, where we have by construction almost no accuracy loss. The average frame rate over the different sequences is near \change{18\,frame/s at this point---an improvement of $9.1\times$ over the cuDNN baseline of 1.96\,frame/s}. 

One sequence (\figref{fig:allEvals}, \ref{trc:gloriastr_highact}) has---while still being close to $5.1\times$ faster than the baseline---a significantly lower throughput than the other sequences. While most of them show typical scenarios such as shown in \figref{fig:gloria_seq}, this sequence shows a very busy situation where the entire road is full of vehicles and all of them are moving. 
\change{The effective number of operations (add or multiply operations) to compute the convolution updates is visualized in \figref{fig:allEvals} (center). For most frame sequences the savings are above $10\times$ while the aforementioned exceptional cases have a significantly higher share with savings of around $5\times$.}

\change{Running the same analysis for the \emph{pose detection} network yields very similar results. For the cuDNN baseline, we get a frame rate of 0.72\,frame/s and CBinfer achieves a rate of 3--8\,frame/s for a threshold factor of 1 or a speed-up of $4.2\times$ to $11.1\times$. A noticeable difference are performance gains for the zero threshold configuration. Here the overhead of CBinfer is outweighed by the savings due to many pixels at the input not changing at all and therefore not triggering an update even for a zero threshold, yielding a performance gain even in a completely loss-less configuration.}

\revision{
\begin{figure}
	\revision{
  \centering
    \begin{tikzpicture} 
        \pgfplotstableread[col sep=comma]
            {\plotdataDirPfm{objDet-sassauna1-eval01-single-baseline.csv}}\dataBaseline;
        \pgfplotstablegetelem{0}{throughput}\of\dataBaseline
        \pgfmathsetmacro{\baselineVal}{\pgfplotsretval}

        \pgfplotstableread[col sep=comma] {\plotdataDirPfm{objDet-sassauna1-eval01-single-seq01.csv}}{\dataTblA}
        \pgfplotstableread[col sep=comma] {\plotdataDirPfm{objDet-sassauna1-eval01-single-seq02.csv}}{\dataTblB}
        \pgfplotstableread[col sep=comma] {\plotdataDirPfm{objDet-sassauna1-eval01-single-seq20.csv}}{\dataTblC}
        \pgfplotstableread[col sep=comma] {\plotdataDirPfm{objDet-sassauna1-eval01-single-seq21.csv}}{\dataTblD}
        \pgfplotstableread[col sep=comma] {\plotdataDirPfm{objDet-sassauna1-eval01-single-seq30.csv}}{\dataTblE}
        \pgfplotstableread[col sep=comma] {\plotdataDirPfm{objDet-sassauna1-eval01-single-seq31.csv}}{\dataTblF}
        \def\objDetPerfTables{
			\dataTblA, 
			\dataTblB, 
			\dataTblC, 
			\dataTblD, 
			\dataTblE, 
			\dataTblF}

        \begin{groupplot}
            [
                group style = {
                    group size=1 by 2,
        xlabels at=edge bottom,
        xticklabels at=edge bottom,
        vertical sep=0pt
                },
                width=0.95\linewidth, 
                height=45mm,
                cycle list name=seqStyleList,
                xmin=0, xmax=2,
			    ylabel absolute, every axis y label/.append style={yshift=-1mm}
            ]
        %%%%% classification error
            \nextgroupplot[
                ylabel={Loss (MSE)}, 
                grid=major, 
                try min ticks=6, 
            ]
            \addplot+ table [x expr=\thisrow{th}/10.0, y=lossIncr] {\dataTblA};
            \addplot+ table [x expr=\thisrow{th}/10.0, y=lossIncr] {\dataTblB};
            \addplot+ table [x expr=\thisrow{th}/10.0, y=lossIncr] {\dataTblC};
            \addplot+ table [x expr=\thisrow{th}/10.0, y=lossIncr] {\dataTblD};
            \addplot+ table [x expr=\thisrow{th}/10.0, y=lossIncr] {\dataTblE};
        %%%%% number of changes
            \nextgroupplot[
                xlabel={Threshold Factor},
                ylabel={\# Operations}, 
                grid=both, 
                try min ticks=6, 
                ymode=log,
               ymin=1e9, ymax=1.2e11
            ]
            \addplot+ table [x expr=\thisrow{th}/10.0, y expr=\thisrow{numGOps}*1e9] {\dataTblA};
            \addplot+ table [x expr=\thisrow{th}/10.0, y expr=\thisrow{numGOps}*1e9] {\dataTblB};
            \addplot+ table [x expr=\thisrow{th}/10.0, y expr=\thisrow{numGOps}*1e9] {\dataTblC};
            \addplot+ table [x expr=\thisrow{th}/10.0, y expr=\thisrow{numGOps}*1e9] {\dataTblD};
            \addplot+ table [x expr=\thisrow{th}/10.0, y expr=\thisrow{numGOps}*1e9] {\dataTblE};
        \end{groupplot}
    \end{tikzpicture}
  \caption{Evaluation of the introduced loss relative to non-change-based inference based on the MSE of the objectness score and the number of executed operations for object detection using YOLOv3. The various traces correspond to different video sequences.}
  \label{fig:perfEvalsObjDet}
  }
\end{figure}
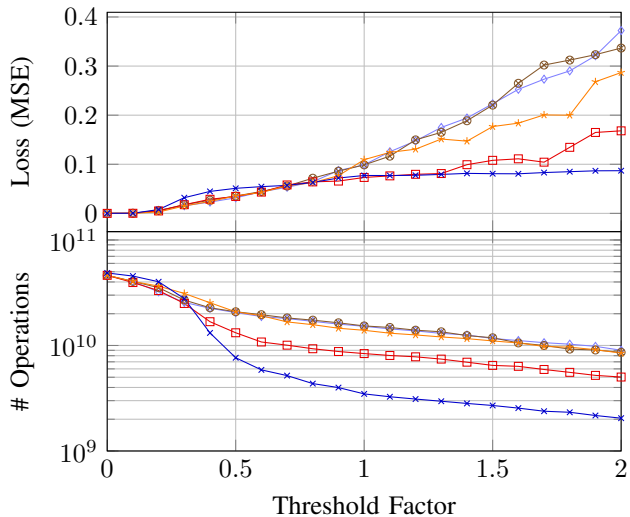
In \figref{fig:perfEvalsObjDet}, we show the evaluation results for \emph{object detection} using YOLOv3 trained on MS COCO and applied to various video sequences. We have observed that the most critical output of the network is the objectness score and that the classification and bounding box dimensions are much more resilient to the change detection threshold. As we do not have ground truth data available for the video sequences used in the experiment, we measure the loss based on the MSE of the objectness score relative to frame-by-frame inference. Again, a clear reduction by around $5\times$ in the number of operations can be observed, albeit not as much as for the other two application scenarios. We attribute this to the network's structure, which uses leaky ReLU activations and hence does not naturally eliminate all changes of feature maps values which are below zero. 
}

\change{We have repeated the performance measurements for the segmentation application with fp32 precision on a workstation with a Nvidia GTX 1080 Ti GPU to compare them to the Tegra X2 platform, obtaining an almost identical throughput-threshold trade-off and compute time breakdown up to a scaling factor of ~$13.9\times$---as can be expected for a largely very well parallelizable workload and a $14.1\times$ more powerful device with a similar architecture\footnote{\change{Tegra X2: 437-750\,GFLOPS (fp32), 874-1500\,GFLOPS (fp16), and 58.4\,GB/s DRAM bandwidth. \\GTX 1080 Ti: 10609\,GFLOPS (fp32) and 484\,GB/s.}}. }

\subsection{Accuracy-Throughput Trade-Off}
\begin{figure}
  \centering
      \begin{tikzpicture}
        \pgfplotstableread[col sep=comma]
            {\plotdataDirPfm{sceneLabeling-brienz2-eval01-single-baseline.csv}}\dataBaseline;
        \pgfplotstablegetelem{0}{throughput}\of\dataBaseline
        \pgfmathsetmacro{\baselineVal}{\pgfplotsretval}
        
        \begin{axis}[
                width=1.0\linewidth,
                height=0.65\linewidth, 
                xlabel={Pixel Classification Accuracy [\%]},
                ylabel={Throughput [frame/s]}, 
                xmin=92.5, xmax=98.5, ymin=0, grid=major, 
                try min ticks=6, 
                cycle list name=seqStyleList,
            ]
        
            \addplot+ table 
                [x=accPixel, y=throughput, col sep=comma]
                {\plotdataDirPfm{sceneLabeling-brienz2-eval01-single-1607__11.csv}};
            \addplot+ table 
                [x=accPixel, y=throughput, col sep=comma]
                {\plotdataDirPfm{sceneLabeling-brienz2-eval01-single-1607__30.csv}};
            \addplot+ table 
                [x=accPixel, y=throughput, col sep=comma]
                {\plotdataDirPfm{sceneLabeling-brienz2-eval01-single-1611__12.csv}};
            \addplot+ table 
                [x=accPixel, y=throughput, col sep=comma]
                {\plotdataDirPfm{sceneLabeling-brienz2-eval01-single-1624__17.csv}};
            \addplot+ table 
                [x=accPixel, y=throughput, col sep=comma]
                {\plotdataDirPfm{sceneLabeling-brienz2-eval01-single-1624__48.csv}};
            \addplot+ table 
                [x=accPixel, y=throughput, col sep=comma]
                {\plotdataDirPfm{sceneLabeling-brienz2-eval01-single-1643__28.csv}};
                
            \draw[thin] (axis cs:\pgfkeysvalueof{/pgfplots/xmin},\baselineVal) -- (axis cs:\pgfkeysvalueof{/pgfplots/xmax},\baselineVal)
            node[above] at (axis cs:{\pgfkeysvalueof{/pgfplots/xmax}-1},\baselineVal) {cuDNN};
    	\end{axis}
    \end{tikzpicture}
    \caption{Evaluation of the throughput-accuracy trade-off \revision{for frame sequences of the Gloriastrasse segmentation dataset. The different frame sequences are marked identically in \figref{fig:allEvals}.}}
    \label{fig:accVSperfAll}
\end{figure}
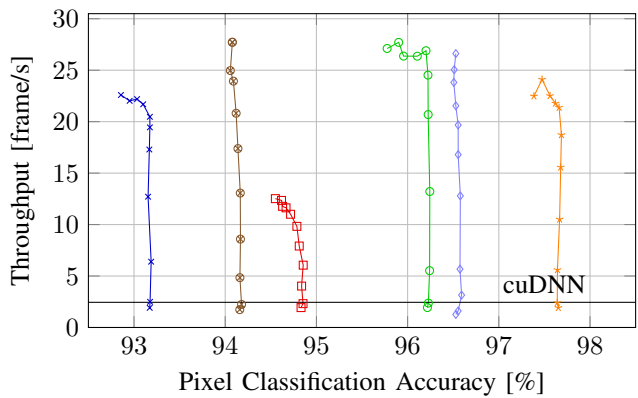
While for some scenarios any drop in accuracy is unacceptable, many applications allow for some trade-off between accuracy and throughput---after all choosing a specific CNN already implies selecting a network with an associated accuracy and computational cost. 

We analyze the trade-off directly in \figref{fig:accVSperfAll}. The most extreme case is updating the entire frame every time resulting in the lowest throughput at the same accuracy as full-frame inference. Increasing the threshold factor in steps of 0.25 immediately results in a significant throughput gain and for most sequences the trade-off only starts at frame rates close to saturation above \change{20\,frame/s}. The same frame sequence that already deviated from the norm before behaves differently here as well. However, an adaptive selection of the threshold factor with a control loop getting feedback about the number of changed pixels could allow for a guaranteed throughput by reducing the accuracy in such cases and is left to be explored in future work.

\subsection{Compute Time Breakdown} \label{sec:compTimeBreakdown}
\begin{figure}
    \centering
    %\tiny
    \scriptsize
    %\small
    \begin{tikzpicture} 
        \begin{axis}[
                xbar stacked,
                symbolic y coords={L3 Conv., L2 Conv., L1 Conv.},
                height=3.8cm, width=\linewidth, xlabel={Compute Time [ms]},
                legend style={at={(0.45,-0.40)}, anchor=north,legend columns=-1}, enlargelimits=0.3, xmin=0, xmax=18.900, enlarge x limits=false,
                try min ticks=9, 
                ytick=data,
                ] 
            
            \addplot coordinates {( 1.990,L1 Conv.) ( 4.490,L2 Conv.) ( 2.819,L3 Conv.)}; 
            \addplot coordinates {( 3.102,L1 Conv.) ( 1.465,L2 Conv.) ( 3.016,L3 Conv.)}; 
            \addplot coordinates {( 2.069,L1 Conv.) ( 1.773,L2 Conv.) ( 1.016,L3 Conv.)}; 
            \addplot coordinates {( 9.341,L1 Conv.) ( 2.256,L2 Conv.) ( 5.490,L3 Conv.)}; 
            \addplot coordinates {( 0.374,L1 Conv.) ( 0.285,L2 Conv.) ( 0.187,L3 Conv.)}; 
            
            \legend{Change Det., Change Extr., gen. X, GEMM, Output Upd.}
            
        \end{axis} 
    \end{tikzpicture}
    \caption{Compute time for the individual processing steps per layer running on the GPU for a typical frame sequence.}
    \label{fig:timebreakdown}
\end{figure}
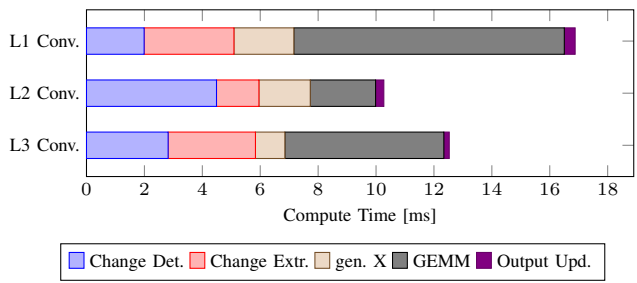
In \secref{sec:baselineComputeBreakdown} and specifically in \tblref{tbl:baselineComputeTime}, we already discussed the compute time breakdown of the entire network when using frame-by-frame analysis. To gain more in-depth understanding of the limiting factors of our proposed algorithm, we show a detailed compute time breakdown of only the change-based convolution layers in \figref{fig:timebreakdown}. The time spent on change detection is similar across all 3 convolution layers, which aligns well with our expectations since the feature map volume at the input of $n_{ch}\cdot h\cdot w$ values is identical for L2 and L3, and 25\% smaller for L1. That this step already makes up for more than 23.4\% of the overall time underlines the importance of a very simple change detection function: any increase in compute time for change detection has to be offset by time savings in the other steps by reducing the number of changes significantly. The change indexes extraction effort is linear to the number of pixels $h\cdot w$ and the clear drop from L1 to L2 is as expected. However, since it is not well parallelizable, there is not much additional gain when comparing L2 to L3. The effort to generate the $\mx{X}$ matrix is very dependent on the number of changed pixels, the number of feature maps, and the filter size. It is, however, most important that the time spent on shuffling data around to generate $\mx{X}$ is significantly smaller than the actual matrix multiplication, which clearly makes up the largest share. The subsequent update of the output values including activation only uses a negligible part of the overall processing time. 

\change{An important aspect is not directly visible: The overall compute time for the dominant convolution layers, has shrunk tremendously by more than $12.9\times$ from \ms{512.8} to about \ms{39.7}. This makes the pooling layer a non-negligible contributor to the overall compute time (total \ms{6.6}). As outlined in \secref{sec:changeProp}, we can perform the pooling also with a change-based approach and skip the change detection and indexes extraction by relying on the preceding convolution layer's change indexes. This provides an additional speed-up by an average of $5.8\times$ and $4.5\times$ for the first and second pooling layer, respectively. 
}
\begin{figure}
	\change{
  \centering
    \begin{tikzpicture} 
    	\begin{axis}[width=\linewidth, 
                height=50mm,
                xmin=0, xmax=2, 
                ymin=0,
                xlabel={Threshold Factor},
                ylabel={\# Operations}, 
                grid=major, 
                area style, stack plots=y, 
                try min ticks=6, ]
            
            \foreach \k in {1, ..., 3} {
            	\addplot+ table [x index=0, y index=\k, col sep=comma] {\plotdataDirPfm{sceneLabeling-eval02-1624__17.csv}} \closedcycle;
            }
            \legend{Layer 1, Layer 2, Layer 3}  	
    	\end{axis}
    \end{tikzpicture}
  \caption{Cumulative number of multiply and add operations for the scene labeling network.}
  \label{fig:OperationsPerLayerSL}
  }
\end{figure}
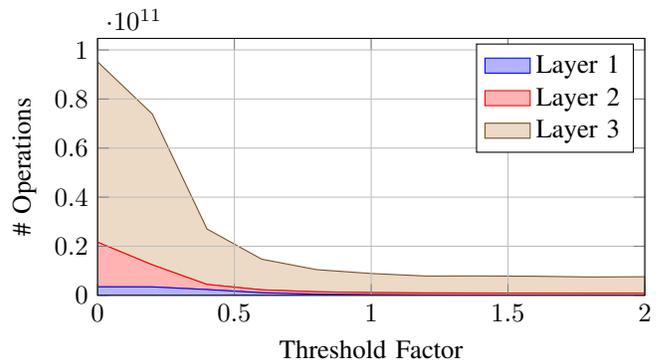
\begin{figure}
	\change{
  \centering
    \begin{tikzpicture} 
    	\begin{axis}[width=\linewidth, 
                height=50mm,
                xmin=0, xmax=2, 
                ymin=0,
                xlabel={Threshold Factor},
                ylabel={\# Operations}, 
                grid=major, 
                area style, stack plots=y, 
                try min ticks=6, ]
            
            \foreach \k in {1, ..., 36} {
            	\addplot+ table [x index=0, y index=\k, col sep=comma] {\plotdataDirPfm{eval02-caviar-walkByShop1cor-seq002-5xSubsample.csv}} \closedcycle;
            }
    	\end{axis}
    \end{tikzpicture}
  \caption{Number of multiply and add operations for the pose detection network stacked by layer with first layer on bottom.}
  \label{fig:OperationsPerLayerPD}
  }
\end{figure}
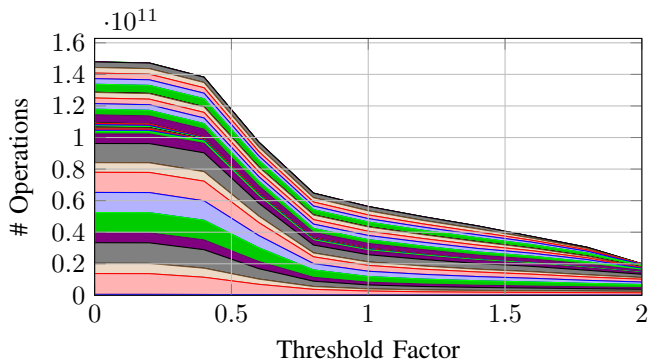

\subsection{Change Propagation}
\change{
During the construction of the algorithm we argued that change detection should be performed for every convolution layer not only for modularity, but also justifying that the worst-case change propagation would result in a rapid growth of the share of changed pixels as we proceed deeper into the network. However, skipping change detection and instead assuming worst-case propagation for some intermediate layers might improve performance. Our experiments have shown that for neither of the networks this pays off. 
An experiment has shown that for Layer~2, the number of changes is reduced by $6.8\times$ from 7.57\% to 1.11\% and for Layer~3 from 2.58\% to 1.94\% by $1.33\times$. Not repeating change detection for some layer affects the compute time: 
\begin{enumerate}
 \item reducing the compute time by substituting the change detection step with a more light-weight change propagation step,
 \item scaling up the compute effort from the matrix generation through the output update proportionally to the increase in the number of pixels marked as change, and
 \item leaving the execution time of the change indexes extraction unaffected.
\end{enumerate}
Combining this with the results in \figref{fig:timebreakdown}, skipping change detection for Layer~2 would result in an increase in execution time by a factor of $2.9\times$. For Layer~3 it would result in approximately no effect on performance.}

\change{An immediate concern evaluating a CNN based on changing pixels is the spreading of the affected regions though the convolutions. We have thus analyzed the effective number of compute operations in \figref{fig:OperationsPerLayerSL} and \figref{fig:OperationsPerLayerPD} for the semantic segmentation and the pose detection networks, respectively. For each layer the number of compute operations is shown in dependence of the joint threshold scaling factor. Layers in parallel branches of the network are shown sequentially. The changes are neither spreading out nor vanishing as we proceed deeper into the DNN. }

\begin{figure}
\change{
    \centering
	\includegraphics[width=\linewidth]{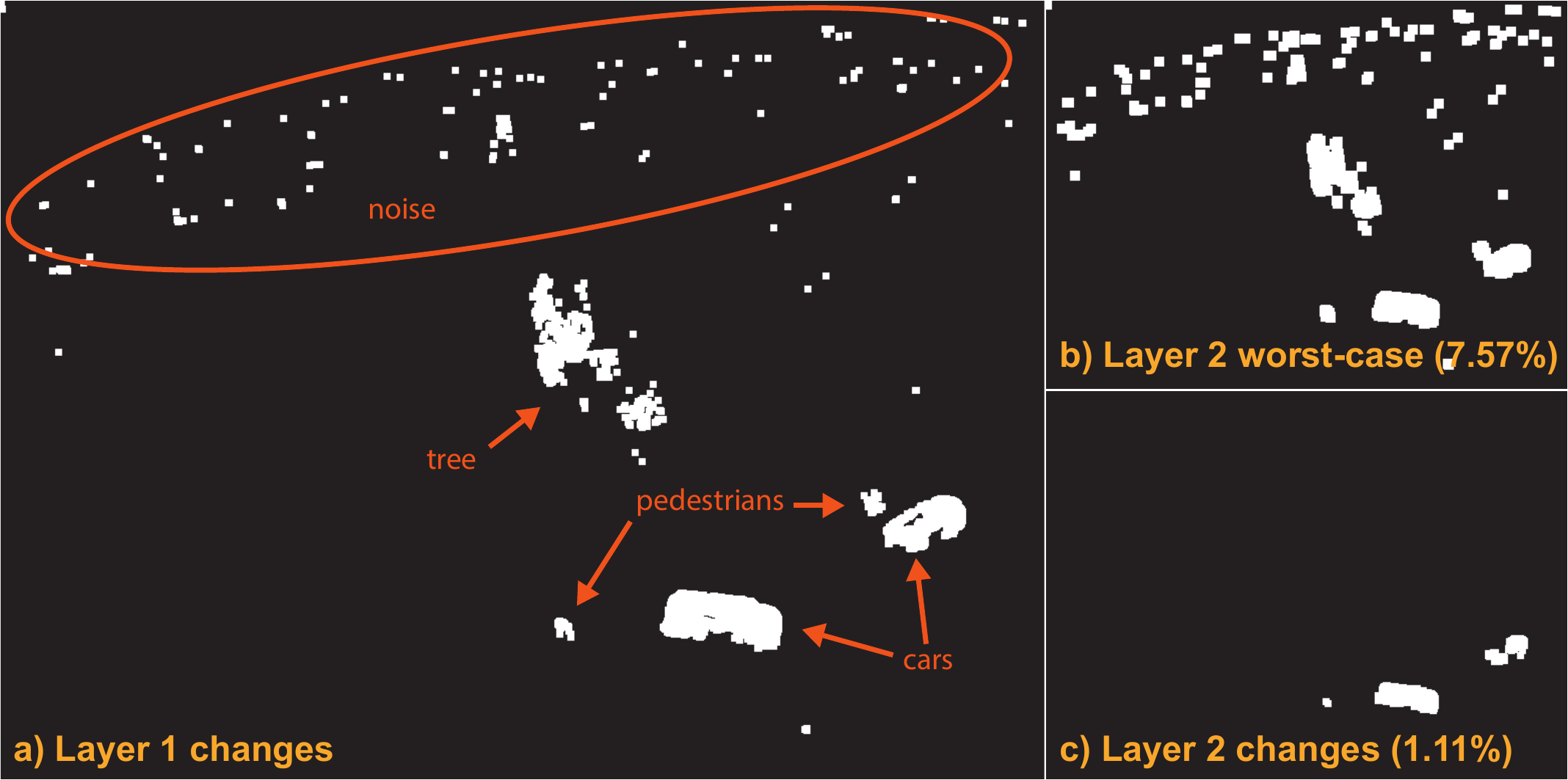}
    \caption{Analysis of the change propagation \revision{with a frame sequence with the same scene as \figref{fig:gloria_seq}}. (a) shows the changes detected \revision{(change map)} in Layer~1 using the thresholds determined in Section~\ref{sec:ThresholdSelResults}, in the upper part of the image there are several single-pixel changes due to noise. We show the changed pixels for Layer~2 based on worst-case propagation as assumed when dropping the Layer~2 change detection step (b) and those when applying change detection instead (c).}
    \label{fig:changeProp}
    }
\end{figure}
\change{
Besides the effect on performance, the visualization of these changes in \figref{fig:changeProp} provides insight into the inner workings of the DNN. As expected, single-pixel artifacts such as noise disappear due to the smoothing effect of the convolution. Changes originating from moving objects such as pedestrians and cars are propagated to the next layer as desired. A particularly interesting observation is the effect on the region marked as tree: In the input frame sequence the leaves of the tree move in the wind, but already after the first convolution layer the resulting changes completely vanish. We construe these pixels in this region to already be represented more abstractly as \emph{leaves}. }

\subsection{Fine-grained CBinfer}
\begin{figure}
  \centering
  \change{
  \revision{
    \begin{tikzpicture} 
        \pgfplotstableread[col sep=comma]
            {\plotdataDirPfm{poseDet-sassauna-eval02b-caviar-walkByShop1cor-seq003-5xSubsample.csv}}\dataTableOpsByLayer;
    	\begin{axis}[width=\linewidth, 
                height=50mm,
                xmin=0, xmax=35, ymin=0,
                xlabel={layer index},
                ylabel={compute effort [GOp]}, 
                grid=major, mark size=1.5pt] {
            	\addplot+ [] table [x=layerIdx, y=normal, col sep=comma] {\dataTableOpsByLayer}; \label{trc:OpPerLayerBL}
            	\addplot+ [] table [x=layerIdx, y=CG, col sep=comma] {\dataTableOpsByLayer}; \label{trc:OpPerLayerCG}
            	\addplot+ [] table [x=layerIdx, y=FG-SP, col sep=comma] {\dataTableOpsByLayer}; \label{trc:OpPerLayerFGSP}
            	\addplot+ [] table [x=layerIdx, y=FG-FM, col sep=comma] {\dataTableOpsByLayer}; \label{trc:OpPerLayerFGFM}
            }	
    	\end{axis}
    \end{tikzpicture}
   }
  \caption{Analysis of the compute effort by layer for the pose detection network. We compare not using CBinfer (\ref{trc:OpPerLayerBL}) to different granularities of CBinfer: normal (coarse-grained) (\ref{trc:OpPerLayerCG}), spatially fine-grained (\ref{trc:OpPerLayerFGSP}), and feature map fine-grained (\ref{trc:OpPerLayerFGFM}).}
  \label{fig:OperationsPerLayerPDdetail}
  }
\end{figure}
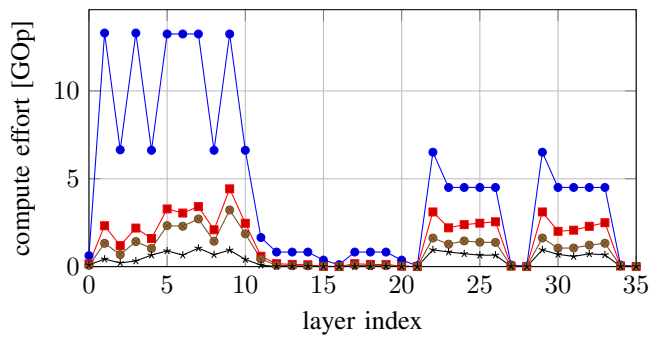

\change{
In \secref{sec:fineGrained} we have introduced two types of fine-grained CBinfer to further reduce the number of multiply-add operations: spatially and across feature maps. 
We analyze this effect by running change-based inference and comparing the compute effort to the number of detected changes and number of operations to perform per change in \figref{fig:OperationsPerLayerPDdetail}. 
The drawbacks discussed in \secref{sec:fineGrained} are confirmed: 
\begin{itemize}
 \item Spatially fine-grained (SP-FG) CBinfer reduces the number of operations only by around 20\% while exploiting this lets the operations become much less regular and the larger matrix multiplication decays into many smaller ones followed by an aggregation step, where both introduce a massive memory bandwidth overhead. 
 \item The results for fine-grained evaluation by feature map show much more potential based on a reduction of multiply and add operations by around 65\%. However, such an implementation also requires a change map per feature map and thus the change extraction step has to be performed on a 3D tensor rather than a 2D tensor. The effort is scaled up by the number of feature maps at the input of the convolution layer (often 16, 64, 256, or more), thereby pushing this computation overhead (for normal CBinfer from 10-20\% of compute time, cf. \figref{fig:timebreakdown}) to several times (160-5120\%) the overall compute effort of normal CBinfer. 
\end{itemize}
}

\subsection{Energy Efficiency}
\begin{figure}
  \centering
    \begin{tikzpicture} 
        \pgfplotstableread[col sep=comma, header=true]
            {\plotdataDirPfm{poseDet-brienz2-eval04-half-recurTrue-pwrModeMAXN-caviar-walkByShop1cor-seq000-5xSubsample.csv}}\datatblA
        \pgfplotstablegetelem{0}{energyBL}\of\datatblA
        \pgfmathsetmacro{\energyBL}{9600}
        \pgfplotstablegetelem{0}{powerBL}\of\datatblA
        \pgfmathsetmacro{\powerBL}{\pgfplotsretval}
        
        \pgfplotstableread[col sep=comma, header=true]
            {\plotdataDirPfm{poseDet-brienz2-eval04-half-recurTrue-pwrModeMAXQ-caviar-walkByShop1cor-seq000-5xSubsample.csv}}\datatblB
        \pgfplotstablegetelem{0}{energyBL}\of\datatblB
        \pgfmathsetmacro{\energyBLmaxQ}{6100}
        \pgfplotstablegetelem{0}{powerBL}\of\datatblB
        \pgfmathsetmacro{\powerBLmaxQ}{\pgfplotsretval}
        
        \pgfplotstableread[col sep=comma, header=has colnames]
            {\plotdataDirPfm{poseDet-brienz2-eval03-half-recurTrue-pwrModeMAXN-caviar-walkByShop1cor-seq000-5xSubsample.csv}}\datatblPerf 
        \pgfplotstablegetelem{0}{execTime}\of\datatblPerf
        \pgfmathsetmacro{\perfBL}{\pgfplotsretval}
        \pgfplotstablegetelem{0}{numGOpsBL}\of\datatblPerf
        \pgfmathsetmacro{\numOpsBL}{\pgfplotsretval}
        
        \pgfplotstableread[col sep=comma, header=has colnames]
            {\plotdataDirPfm{poseDet-brienz2-eval03-half-recurTrue-pwrModeMAXQ-caviar-walkByShop1cor-seq000-5xSubsample.csv}}\datatblPerfMaxq
        \pgfplotstablegetelem{0}{execTime}\of\datatblPerfMaxq
        \pgfmathsetmacro{\perfBLMaxq}{\pgfplotsretval}
        
    	\begin{groupplot}[
    			group style={group size=1 by 5, xlabels at=edge bottom,xticklabels at=edge bottom,vertical sep=3pt},
    			width=\linewidth, 
                height=40mm,
                xmin=0, xmax=200, ymin=0, 
                xlabel={time (frame index)},                
                grid=major, ]
    	\nextgroupplot[ylabel={Loss}, ymax=2.2e-3]
            \addplot[red] table [x=frameIdx, y expr=1e-3*\thisrow{loss}] {\datatblPerf};
            
    	\nextgroupplot[ylabel={Runtime [s/frm]}]
            \addplot[red] table [x=frameIdx, y=execTime] {\datatblPerf};
            \addplot[red, dashed] table [x=frameIdx, y=execTime] {\datatblPerfMaxq};
            \draw[blue] (axis cs:\pgfkeysvalueof{/pgfplots/xmin},\perfBL) -- (axis cs:\pgfkeysvalueof{/pgfplots/xmax},\perfBL);
            \draw[blue, dashed] (axis cs:\pgfkeysvalueof{/pgfplots/xmin},\perfBLMaxq) -- (axis cs:\pgfkeysvalueof{/pgfplots/xmax},\perfBLMaxq);
            
    	\nextgroupplot[ylabel={GOp/frame}, ymax=18e1]
            \addplot[red] table [x=frameIdx, y=numGOps] {\datatblPerf};
            \draw[blue] (axis cs:\pgfkeysvalueof{/pgfplots/xmin},\numOpsBL) -- (axis cs:\pgfkeysvalueof{/pgfplots/xmax},\numOpsBL);
            
    	\nextgroupplot[ylabel={Power [W]},ymax=14,]
            \addplot[red] table [x=frameIdx, y expr=\thisrow{power}/1000.0] {\datatblA};
            \addplot[red, dashed] table [x=frameIdx, y expr=\thisrow{power}/1000.0] {\datatblB};
            \draw[blue] (axis cs:\pgfkeysvalueof{/pgfplots/xmin},\powerBL/1000.0) -- (axis cs:\pgfkeysvalueof{/pgfplots/xmax},\powerBL/1000.0);
            \draw[blue, dashed] (axis cs:\pgfkeysvalueof{/pgfplots/xmin},\powerBLmaxQ/1000.0) -- (axis cs:\pgfkeysvalueof{/pgfplots/xmax},\powerBLmaxQ/1000.0);
            
    	\nextgroupplot[ylabel={Energy [mJ/frm]},ymax=9.99]
    		\addplot[red, mark=x] table [x=frameIdx, y expr=1e-3*\thisrow{energy}] {\datatblA}; \label{trc:energyCbinferMaxn}
            \addplot[dashed, red, mark=x] table [x=frameIdx, y expr=1e-3*\thisrow{energy}] {\datatblB}; \label{trc:energyCbinferMaxq}
            \draw[blue] (axis cs:\pgfkeysvalueof{/pgfplots/xmin},1e-3*\energyBL) -- (axis cs:\pgfkeysvalueof{/pgfplots/xmax},1e-3*\energyBL); 
            \addplot[blue] coordinates {(-1,-1)};\label{trc:energyCudnnMaxn}
            \draw[blue, dashed] (axis cs:\pgfkeysvalueof{/pgfplots/xmin},1e-3*\energyBLmaxQ) -- (axis cs:\pgfkeysvalueof{/pgfplots/xmax},1e-3*\energyBLmaxQ); 
            \addplot[blue, dashed] coordinates {(-1,-1)}; \label{trc:energyCudnnMaxq}
    	\end{groupplot}
    \end{tikzpicture}
  \caption{\change{Evaluation of accuracy effect, runtime, number of compute operations, power and energy when performing inference for the pose detection network on a 200-frame video sequence.} \textbf{Legend:} {\color{red}CBinfer}, {\color{blue}cuDNN}; continuous: max-N (max. performance) power mode, dashed: max-Q (max. efficiency) power mode.}
  \label{fig:timeTrace}
\end{figure}
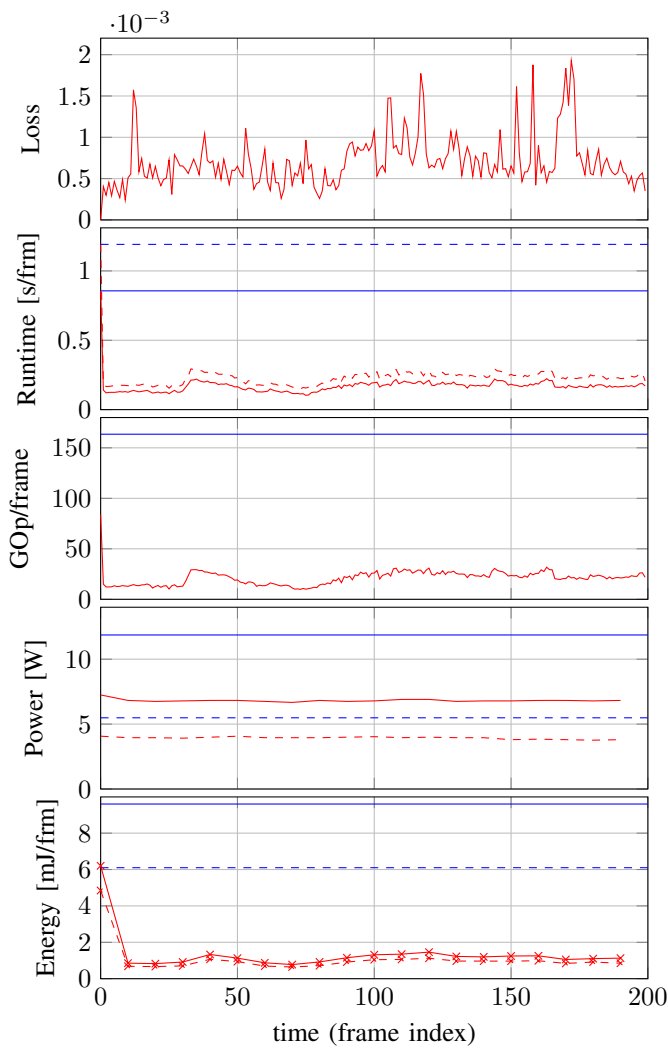
\change{
We have measured the power consumption of the Tegra X2 module using the on-board sensors for two of its power modes: maximum performance (max-N) and maximum efficiency (max-Q). When idling, the power consumption is 1.80\,W and 1.77\,W for max-N and max-Q, respectively. The measurements under load have been conducted while running pose detection on a 200-frame sequence and are visualized in \figref{fig:timeTrace}. 
Generally, we can see a clear correlation between the number of operations that have to be computed and the runtime, where the later has a clear offset due to the overhead of change detection and change indexes extraction. We can also observe that there is no long-term rise in the introduced loss. The power is very constant for the cuDNN baseline in max-N (\ref{trc:energyCudnnMaxn}, 12\,W) and max-Q mode (\ref{trc:energyCudnnMaxq}, 5.3\,W) as well as for the CBinfer implementation (\ref{trc:energyCbinferMaxn}, 6.8\,W) and (\ref{trc:energyCbinferMaxq}, 4.8\,W), respectively. Note that we were processing the frames without duty-cycling. The resulting energy efficiency is shown in the trace at the bottom. The baseline uses around 9.6\,J/frame in max-N mode and 6.1\,J/frame in max-Q mode, whereas the CBinfer implementation uses an average of 1.1\,J/frame and 0.8\,J/frame, respectively. This corresponds to energy savings of $8.7\times$ and $7.6\times$ and an equivalent average energy efficiency of 148\,GOp/s/W and 204\,GOp/s/W for the max-N and max-Q power modes, respectively. 

For the scene labeling network and the max-N power mode we have measured a power consumption of 6.8\,W with CBinfer and 10.5\,W with cuDNN and thus 411 and 3003\,mJ/frame, respectively. With a frame requiring 210\,GOp, this results in an energy efficiency of 511 and 70\,GOp/s/W---an improvement by $5.9\times$.
}
\revision{
\subsection{Comparison to Related Work}
\revisionB{
CBinfer is compared to state-of-the-art methods exploiting temporal redundancy in \tblref{tbl:resultComparison}. DeepMon \cite{Huynh2017} achieves a speed-up of 46\% using VGG-16 on the UCF101 dataset at an accuracy drop of 6\% from 89.9\% to 83.9\%. CNNCache \cite{Xu2017} improves on this result by achieving an 23\% speed-up at an accuracy loss of 3\% using ResNet-50 on the simplified UCF101 dataset with 10 classes. 

With CBinfer, the average speed-up is in the range of 700--910\% at negligible accuracy loss (e.g. $<0.05\%$ on average for segmentation). These results should not be compared directly with the previous methods: since the video sequences of the UCF101 dataset have a moving camera and we require a static camera, the evaluation had to be performed on different datasets. }

\begin{table*}
\revisionB{
	\caption{Comparison of Results with State-of-the-Art}
	\label{tbl:resultComparison}
	\begin{tabularx}{\linewidth}{lXlX}
		\toprule\parnoteclear 
		Method & Dataset & Camera/Backgr. & Result (speed-up @ accuracy loss) \\ \midrule
		DeepMon & UCF101 simplified (activity/sports recordings, 10 cl.) & object-following & 1.23$\times$ speed-up @ 3\% loss with ResNet-50\\
		CNNCache & UCF101 (activity/sports recordings, 101 classes) & object-following & $1.46\times$ speed-up @ 6\% loss with VGG-16\parnote{Only speed-up from using convolutional layer caching. Additional methods are also presented in the corresponding work.}\\
		CBinfer (ours) & Gloriastr. (surveillance cam, semantic segm., 10 classes) & static & 9.1$\times$ speed-up @ $<$0.05\% loss\\
		CBinfer (ours) & CAVIAR video sequences (for pose detection) & static & 7.0$\times$ speed-up @ negligible loss with OpenPose\parnote{Difference to non-change-based inference of less than $0.3\cdot 10^{-3}$ MSE.}\\
		\bottomrule
	\end{tabularx}
	\parnotes
}
\end{table*}

However, we can discuss the origins of the limited speed-ups and how they are overcome using our method. The static camera requirement allows us to eliminate the costly image block matching employed by CNNCache to find the corresponding image block in the previous frame. In our case, the matching becomes a trivial per-pixel comparison, since it is immediately clear where to find the corresponding patch of pixels in the previous frame. CNNCache hence also has to limit the identification of where cached results can be re-used to the input of the CNN, since repeated block matching with many feature maps would introduce an overhead likely in excess of the compute time for the CNN itself. 

The repetition of the change detection at each layer is a crucial property of our algorithm and allows us to reuse significantly more data from the cache (previous frame's output of each layer) because we are not bound to fetch rectangular regions from cache and can eliminate irrelevant changes at an early stage within the network. This is particularly important for our application scenarios, where we also expect multiple small moving objects of interest in the scene while expecting changing pixels from background object irrelevant to the final output. 
}

\section{Conclusion} \label{sec:conclusion}
\change{
We have proposed and evaluated a novel algorithm for change-based evaluation of CNNs for video recorded with a static camera setting, exploiting the spatio-\allowbreak temporal sparsity of pixel changes. 
The method introduces a set of parameters to trade-off accuracy and throughput. Even when choosing the parameters conservatively to introduce no significant accuracy loss, we have observed an average speed-up by $9.1\times$ for a semantic segmentation DNN and $7.0\times$ for a pose detection DNN relative to cuDNN using our GPU implementation.
The resulting boost in energy efficiency over per-frame evaluation is an average of $8.7\times$ and $5.9\times$ for the two applications, respectively. This corresponds to an equivalent energy efficiency of 511\,GOp/s/W on the Tegra~X2 platform for the pose detection DNN. For object detection using YOLOv3, we have observed a reduction of $5\times$ in computational workload. 
We have furthermore analyzed various flavors of the proposed algorithm and how the changes propagate through the DNNs to further underline the optimality of the structure of the proposed algorithm. 

Further gains might be possible by training the network on videos using change-based inference for the forward propagation or by introducing noise to simulate the slight deviations for the ideal feature maps. The proposed method should also not be limited to video data, but work on any data where changes in at least one dimension are sparse (e.g. spectrograms of audio data). Finally, reducing the granularity of the algorithm by $2\times 2$ or $4\times 4$ would allow an implementation using Winograd's convolution algorithm for additional speed-up, like it is being done in the cuDNN baseline. 
}

\bibliographystyle{IEEEtran}
\bibliography{library}

\end{document}